\documentclass{article} 
\PassOptionsToPackage{table}{xcolor}
\usepackage{iclr2026_conference,times}

\usepackage{amsmath,amsfonts,bm}

\def\eqref#1{equation~\ref{#1}}

\def\1{\bm{1}}

\DeclareMathAlphabet{\mathsfit}{\encodingdefault}{\sfdefault}{m}{sl}
\SetMathAlphabet{\mathsfit}{bold}{\encodingdefault}{\sfdefault}{bx}{n}

\newcommand{\E}{\mathbb{E}}

\DeclareMathOperator*{\argmin}{arg\,min}

\usepackage{threeparttable}
\usepackage{tabularx}
\usepackage{arydshln}
\usepackage{booktabs}
\usepackage{times}
\usepackage{epsfig}
\usepackage{graphicx}
\usepackage{amsmath}
\usepackage{amssymb}
\usepackage[english]{babel}
\usepackage{pdfrender}

\definecolor{citepink}{RGB}{185,80,135}
\definecolor{NavySoft}{HTML}{3A5A78}
\usepackage[colorlinks=true,
            citecolor=NavySoft]{hyperref}
\hypersetup{colorlinks}
\usepackage{cleveref}

\usepackage{url}
\usepackage{mathtools}
\usepackage{dpr}
\usepackage{fec}
\usepackage{enumerate}
\usepackage{subcaption}
\usepackage{algorithm,algpseudocode}
\usepackage{makecell}
\usepackage{comment}
\usepackage{caption}
\usepackage{subcaption}
\usepackage{enumitem}
\usepackage{amsthm}
\usepackage{bm}
\usepackage{siunitx,tabularx,ragged2e} 
\usepackage{wrapfig}
\usepackage{float}
\usepackage{pifont}
\usepackage{soul}
\usepackage{epstopdf}
\usepackage{blindtext}
\usepackage{fancyvrb}
\usepackage{multirow, booktabs}
\usepackage{hhline}
\usepackage{textcomp}
\usepackage{tcolorbox}
\usepackage{MnSymbol}
\usepackage{amssymb,fge}
\usepackage{setspace}
\usepackage{subcaption}
\usepackage{bbding}
\usepackage{hhline}
\usepackage{natbib}
\usepackage{adjustbox}

\setcitestyle{authoryear,open={(},close={)}} 

\usepackage[table]{xcolor}

\definecolor{cPeach}{HTML}{E78160}
\definecolor{cSky}{HTML}{56B4E9}
\definecolor{cAmber}{HTML}{E69F00}
\definecolor{cMagenta}{HTML}{AA4499}

\definecolor{tabcoral}{HTML}{E78160}   
\definecolor{tabskyblue}{HTML}{56B4E9} 
\colorlet{tabcoral}{tabcoral}
\colorlet{tabskyblue}{tabskyblue}
\colorlet{cAmber}{cAmber}
\colorlet{cMagenta}{cMagenta}

\definecolor{goodgreen}{HTML}{009E73} 
\definecolor{darkpastelgreen}{rgb}{0.01, 0.75, 0.24}
\definecolor{electriccrimson}{rgb}{1.0, 0.0, 0.25}
\definecolor{navyblue}{rgb}{0.0, 0.0, 0.75}
\newcommand{\ie}{\text{i.e.}}
\newcommand{\eg}{\text{e.g.}}

\theoremstyle{plain}
\newtheorem{theorem}{Theorem}[section]

\newtheorem{lemma}[theorem]{Lemma}
\newtheorem{corollary}[theorem]{Corollary}
\theoremstyle{definition}
\newtheorem{definition}[theorem]{Definition}

\newcommand{\algacro}{WINA{}}

\newcommand{\cmark}{\textcolor{green!70!black}{\small\ding{51}}}
\newcommand{\xmark}{\textcolor{red}{\small\ding{55}}}

\newcolumntype{C}{>{\centering\arraybackslash}X}

\title{\algacro{}: Weight Informed Neuron Activation for Accelerating Large Language Model Inference}

\author{
Sihan Chen$^{2*}$, Dan Zhao$^{1}$, Jongwoo Ko$^1$, Colby Banbury$^1$, Huiping Zhuang$^3$, \\ \ \textbf{Luming Liang$^1$, Pashmina Cameron$^1$, Tianyi Chen$^{1\dagger}$\thanks{Equal contributions. \dagger Corresponding author.}} \\ 
$^1$Microsoft, $^2$Renmin University of China, $^3$South China University of Technology\\
\ \small{\href{mailto:chensihan@ruc.edu.cn}{\texttt{chensihan@ruc.edu.cn}},
\href{mailto:tiachen@microsoft.com}{\texttt{Tianyi.Chen@microsoft.com}}}
}

\iclrfinalcopy 
\begin{document}

\maketitle

\begin{abstract}
The ever-increasing computational demands of large language models (LLMs) make efficient inference a central challenge. While recent advances leverage specialized architectures or selective activation, they typically require (re)training or architectural modifications, limiting their broad applicability. Training-free sparse activation, in contrast, offers a plug-and-play pathway to efficiency; however, existing methods often rely solely on hidden state magnitudes, leading to significant approximation error and performance degradation. To address this, we introduce \textbf{\algacro{}} (\textbf{W}eight-\textbf{I}nformed \textbf{N}euron \textbf{A}ctivation): a simple framework for training-free sparse activation that incorporates both hidden state magnitudes and weight matrix structure. By also leveraging the $\ell_2$-norm of the model's weight matrices, \algacro{} yields a principled sparsification strategy with provably optimal approximation error bounds, offering better and tighter theoretical guarantees than prior state-of-the-art approaches. Overall, \algacro{} also empirically outperforms many previous training-free methods across diverse LLM architectures and datasets: not only matching or exceeding their accuracy at comparable sparsity levels, but also sustaining performance better at more extreme sparsity levels. Together, these position \algacro{} as a practical, theoretically grounded, and broadly deployable solution for efficient inference. Our code is available at \href{https://github.com/microsoft/wina}{github.com/microsoft/wina}.
\end{abstract}

\section{Introduction}

While large language models (LLMs) have demonstrated impressive capabilities in a variety of applications such as text generation~\citep{li2024pre,cheng2025dehumanizing}, translation~\citep{hendy2023goodgptmodelsmachine,seamless2025joint},  understanding~\citep{chang2024survey,tschannen2025siglip}, and computer using agent \citep{xie2024osworld,hui2025winclick}. Their growing size and complexity often translate into a need for substantial computational resources, particularly during inference, making reducing inference costs without degrading output quality a key challenge.

One strategy is to only activate a sub-network~\citep{jacobs1991adaptive} during inference through architectural changes such as Mixture of Experts (MoE), which has already seen adoption in LLMs like GPT4~\citep{achiam2023gpt}, Mistral~\citep{jiang2023mistral}, etc., or through model distillation~{\citep{moslemi2024survey}} where a smaller model is trained using knowledge distilled from a larger teacher model to route inference requests more efficiently. However, these approaches typically require a considerable amount of training which, in itself, can be computationally intense.

One alternative is training-free sparse activation~{\citep{liu2024trainingfreeactivationsparsitylarge,lee2024catscontextuallyawarethresholdingsparsity}} which retains the original dense model but selectively omits weights/neurons at inference time. These methods avoid (re)training and can be applied to off-the-shelf models, leveraging criteria such as hidden-state magnitudes, weight importance, weight statistics, or additional validation data to determine which parts of the model to select to accelerate inference.

Nonetheless, current training-free sparse-methods face several limitations. Most notably, they ignore the influence of weight matrices on error propagation: these approaches fail to account for how interactions between input elements and the weight matrix during forward propagation can affect model outputs, which lead to accumulating approximation errors in sparse activation.

\paragraph{Contributions.} We propose \algacro{}: a simple, easy-to-use, training-free framework that performs sparse activation based on both the magnitude of hidden states and the column-wise $\ell_2$-norm of the weight matrix. By combining activation strength with weight importance, our thresholds directly account for how much each activation can influence the next layer. This design provides theoretical guarantees on bounding the total approximation error in a way better than that of other approaches. 

In contrast, methods like TEAL~\citep{liu2024trainingfreeactivationsparsitylarge} rely exclusively on the distribution of hidden‐state magnitudes to decide which activations to keep. However, ignoring weight magnitudes in this way can discard highly influential activations while retaining many low‐impact ones, leading to suboptimal trade-offs between efficiency and output quality. 
These types of approaches overlook how the weight matrix directly influences input features during the forward pass; by focusing solely on activations, this
omission can result in compounding approximation errors that ultimately skew the calculation of sparse activations and degrade the accuracy of the sparse model more than necessary. Our framework overcomes these limitations by integrating weight statistics into the selection process, achieving finer control over sparsity and tighter bounds on the resulting approximation error.

\begin{wraptable}{r}{0.5\textwidth}
\vspace{-3mm}
\centering
\caption{Comparison between \algacro{} and others.}
\vspace{-2mm}
\scriptsize
\resizebox{\linewidth}{!}{%
\begin{tabular}{lcccc}
\toprule
 & \textbf{\algacro{}} & \textbf{TEAL}  & \textbf{CATS} & \textbf{R-Sparse}  \\
\midrule
\textbf{Tight Approx Error}  & \cellcolor{yellow!20}{\cmark} & \xmark & \xmark & \xmark \\
\textbf{Layer-Agnostic Application}$^\dagger$ & \cellcolor{yellow!20}{\cmark} & \cmark & \xmark & \cmark\\
\textbf{Layer-Specific Sparsity} & \cellcolor{yellow!20}{\cmark} & \cmark & \xmark & \cmark  \\
\bottomrule
\end{tabular}}
{\fontsize{6}{10}\selectfont
\begin{tabular}{@{}p{0.9\textwidth}@{}}
$\dagger$: Some methods (i.e., CATS) are only adopted on specific types of layers. \\
\end{tabular}
}
\vspace{-5mm}
\end{wraptable}

We evaluate \algacro{} on several widely-used LLMs (ranging from 7B to 14B) across several popular benchmarks. Compared with existing training-free sparse activation methods like TEAL~\citep{liu2024trainingfreeactivationsparsitylarge}, CATS~\citep{lee2024catscontextuallyawarethresholdingsparsity}, and R-Sparse~\citep{zhang2025r}, \algacro{} achieves superior model performance at comparable sparsity levels with significantly less performance degradation. 
We also establish tight theoretical error bounds for \algacro{},  providing formal support for our experimental results to validate its effectiveness. In short, our contributions can be summarized as follows:

\begin{itemize}[leftmargin=*, itemsep=3pt]
    \item \textbf{Weighted-Informed Activation.} We introduce a novel sparse activation method that jointly considers hidden state magnitudes and the column-wise $\ell_2$-norms of weight matrices. This allows for selecting neurons that are not only strongly activated but also those that have a larger influence on downstream layers, leading to a more informed construction of a sub-network during inference.
    \item \textbf{Theoretically Tighter Approximation Error.} {We conduct a formal analysis to demonstrate that our weight-informed activation mechanism yields a lower expected output error compared to prior methods (e.g., TEAL) under mild assumptions.}
    \item \textbf{Numerical Experiments.} {We conduct extensive evaluations across multiple LLM families, including the Lllama series~\citep{touvron2023llama}, Phi-4~\citep{abdin2024phi}, and Mistral~\citep{jiang2023mistral}, and show that our method consistently achieves superior accuracy across a range of sparsity levels. In particular, \algacro{} sustains higher performance as sparsity increases, highlighting its robustness and practical applicability across diverse tasks and model scale. Moreover, we also demonstrate \algacro{}'s compatibility with quantization, achieving promising results under both 4-bit and 8-bit settings, \algacro{}'s impact on models' long-context reasoning abilities, and its impact on social biases in models---aspects largely overlooked in prior works. Lastly, we provide a Triton kernel to benchmark \algacro{}'s competitive speed-ups against methods like TEAL.}
\end{itemize}

The rest of our paper is organized as follows. We begin with related works in Section \ref{sec:related_work} and detail our methodology in Section \ref{sec:algo}. We present our results in Section \ref{sec:experiments} and conclude in Section \ref{sec:conclusion}.

\section{Related Work}
\label{sec:related_work}

\paragraph{Sparse Activation.} Modern sparse activation approaches fall into two principal paradigms: training-based methods and training-free methods. Training-based methods typically employ a trainable router to learn to dynamically select activated experts for each token, with the Mixture-of-Experts (MoE) architecture~\citep{jacobs1991adaptive} serving as the foundational framework.

This paradigm has been expanded through many iterations and variants. The sparsely-gated mixture of experts layer \citep{shazeer2017outrageously} integrates MoE into recurring neural networks (RNNs). Works like GShard \citep{lepikhin2020GShard} and the Switch Transformer \citep{fedus2022switch} extend MoEs to the Transformer architecture \citep{raffel2020exploring} while others combine several approaches, such as WideNet~\citep{xue2022go},  reduces the size of the MoE model by initially compressing the model before transitioning into a MoE. Works like MoEBert \citep{zuo2022moebert} decomposes the FFN layer of a pre-trained dense model into multiple experts based on importance-guided adaptation and then refines the model through distillation. LLM in Flash \citep{alizadeh2023llm} employs a low-rank predictor to determine which intermediate neurons are activated.

Training-free methods, in contrast, do not rely on a learnable router, instead using predefined criteria to perform sparse activation such as TEAL \citep{liu2024trainingfreeactivationsparsitylarge}, TDA \citep{ma2024activationsmattertrainingfreemethods}, and SEAP \citep{liang2025seaptrainingfreesparseexpert}. Methods \citep{han2015learnweightsconnect} can utilize magnitude-based weight pruning or global activation pruning \citep{wen2016learnstructsparse} to apply a fixed sparsity pattern regardless of input. For instance, Q-Sparse \citep{wang2024q} produces sparsity as a function of input magnitudes, achieving high sparsity with reasonable performance degradation. CATS~\citep{lee2024catscontextuallyawarethresholdingsparsity} applies sparse activation on SwiGLU within gated MLP layers, achieving reasonable sparsity while maintaining performance. In contrast, TEAL~\citep{liu2024trainingfreeactivationsparsitylarge} extends magnitude-based activation sparsity to all layers, achieving high model-wide sparsity. However, current sparse activation methods tend to focus solely on selecting activation elements based on the magnitude of hidden states, which can result in suboptimal performance.

\vspace{-3mm}
\paragraph{Model Pruning.} 
Another line of related work is model pruning, which removes redundant neurons from deep neural networks~\citep{han2015deep, frankle2018lottery, frantar2023sparsegpt,li2024greedyoutputapproximationefficient} to obtain compact yet high-performing sub-networks. While effective, pruning typically requires fine-tuning to recover accuracy~\citep{lin2019toward, he2018soft, wen2016learning, li2020group, zhuang2020neuron, chen2021oto, chen2024hesso}, introducing additional training overhead. Such retraining stages make pruning less practical for large foundation models.
\section{Methodology}
\label{sec:algo}

\begin{figure}
    \centering
    \includegraphics[width=0.95\textwidth]{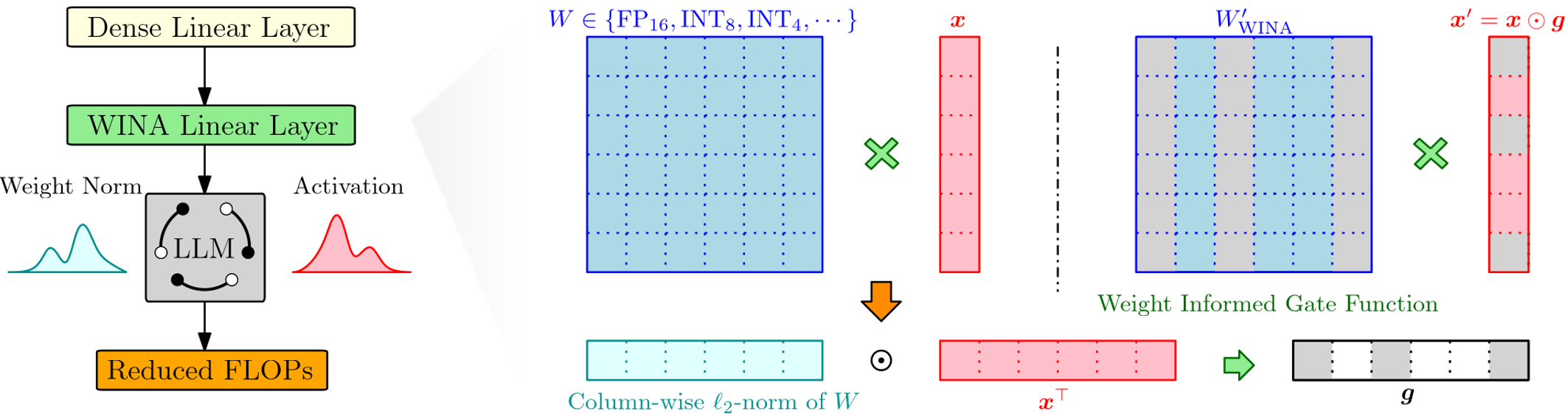}
    \caption{{\textbf{Overview of \algacro{}.} \algacro{} performs training-free sparse activation by selecting the most influential input dimensions based on both activation magnitude and the column-wise $\ell_2$-norms of weight matrices. This joint criterion effectively extends to layers of varying precision, including quantized layers, ensuring accurate sub-network activation at each layer during inference. 
    }}
    \label{fig:Overview}
\end{figure}

We now present \algacro{}, a framework for sparse activation that preserves critical elements while zeroing out non-essential components in each layer's input. As illustrated in Figure~\ref{fig:Overview}, \algacro{} jointly considers both the input  activation and the associated weight matrix, rather than relying solely on activation magnitudes. During inference, it activates only the most influential neurons, effectively constructing a sparse sub-network that maintains the expressive power of the original model.

\subsection{Problem Statement}

{
\textbf{Problem.}  
Consider a deep neural network (DNN) $\mathcal{M}$ consisting of $L$ layers. We denote the weight matrix of the $l$-th layer as $W^{(l)} \in \mathbb{R}^{m_l \times n_l}$ and the corresponding input as an arbitrary tensor $X \in \mathbb{R}^{n_l \times s_l}$ for $l \in \{1, ..., L\}$, representing the full information content.  
Our goal is to identify a set of binary activation gates $\mathcal{G} = \{\bm{g}^{(1)}, \cdots, \bm{g}^{(L)}\}$, where each $\bm{g}^{(l)} \in \{0, 1\}^{n_l}$ is a binary vector, such that the deviation between the model's original output and the gated output is minimized:
\begin{equation}
\mathop{\text{minimize}}_{\bm{g}^{(1)}, \cdots, \bm{g}^{(L)}} \quad \norm{\mathcal{M}(X) - \mathcal{M}(X \mid \mathcal{G})}_2.
\end{equation}
Since obtaining the complete set of possible inputs $X$ is generally infeasible, we instead use a sampled subset $\tilde{X}$ for approximation. 
The activation gating operates in the input vector space to reduce output deviation. With this observation, we can reformulate the original problem into a more tractable per-layer version.

\textbf{Refined Problem.}  
Given a weight matrix $W \in \mathbb{R}^{m \times n}$ and input vector $\bm{x} \in \mathbb{R}^{n}$, the standard linear transformation is given by $\bm{y} \gets W \bm{x}$.  
Our objective then becomes identifying an activation gate or mask $\bm{g} \in \{0, 1\}^{n}$ such that the masked output $\bm{y}_{\bm{g}} \gets W(\bm{g} \odot \bm{x})$ approximates the original:
\begin{equation}\label{main.prob}
\mathop{\text{minimize}}_{\bm{g} \in \{0, 1\}^{n}} \quad  \norm{W \bm{x} - W(\bm{g} \odot \bm{x})}_2.
\end{equation}
}
\subsection{Weight Informed Gate Function}

\textbf{Motivation.}
Many current sparse activation methods (e.g., Q-sparse~\citep{wang2024q}, CATS~\citep{lee2024catscontextuallyawarethresholdingsparsity}, TEAL~\citep{liu2024trainingfreeactivationsparsitylarge}) operate via a top-$K$ gating mechanism governed by the absolute values of the hidden states:
\begin{equation}\label{eq:wina_gate}
\bm{g}_i = \begin{cases}
1 & \text{if } |\bm{x}_i| \text{ is among the top-$K$ values in } |\bm{x}|, \\
0 & \text{otherwise}.
\end{cases}
\end{equation}
However, this approach ignores the critical role that weight matrices themselves play: more specifically, it ignores how each element of the preceding input interacts with the weight matrix $W$ during the forward pass. This mismatch is what motivates us to propose \algacro{}, a method that jointly considers both inputs and weight matrices to minimize the approximation error for better performance.

In \algacro{}, we construct binary activation gates by selecting the top-$K$ components via:
\begin{equation}\label{eq:wina_gate}
[\bm{g}_{\text{\algacro{}}}]_i = \begin{cases}
1 & \text{if } |\bm{x}_i \bm{c}_i| \text{ is among the top-$K$ values in } |\bm{x} \odot \bm{c}|, \\
0 & \text{otherwise},
\end{cases}
\end{equation}
where $\bm{c} \in \mathbb{R}^n$ represents the column-wise $\ell_2$-norm of $W$ and $\odot$ denotes the Hadamard or element-wise product. A smaller $K$ results in more deactivated neurons, thereby saving more FLOPs while potentially sacrificing performance. The choice of $K$ is flexible and adaptable, ranging from a coarse-grained universal criterion where a shared $K$ is applied across all layers to a fine-grained layer-specific strategy that assigns $K$ individually to better minimize performance degradation. In the meantime, \algacro{} is architecture agnostic, applicable across different layers, such as attention layers, multi-layer perceptions (MLPs), or residual connections, etc.

\subsection{Theoretically Optimal Approximation Error}

\algacro{} also comes with theoretical guarantees, establishing a tighter bound on the approximation error than prior approaches under relatively mild assumptions. We first show that for a single linear layer network, \algacro{} yields an optimal solution to the target problem~(Eqn. \ref{main.prob}).

\begin{lemma}[Optimal approximation error over single linear layer]\label{lemma.single_layer}
Let  $\bm{x} \in \mathbb{R}^n$ be an input vector and $W \in \mathbb{R}^{m \times n}$ be a matrix satisfying column-wise orthogonality: $W^{\top}W$ is a diagonal matrix.
For any target density level $K \in \mathbb{N}^+$ satisfying $K < n$, 
the deviation between the original network output and the gated output via \textit{\algacro{}} is optimal. Formally, 
$$
\bm{g}_{\text{\algacro{}}}(\bm{x})=\argmin_{\bm{g}\in\{0,1\}^{n}}\quad \norm{W\bm{x} - W(\bm{g}\odot \bm{x})}_2,
$$
where $\bm{x}$ is the input, and $\bm{g}_{\text{\algacro{}}}(\bm{x})$ is the gating function of~\algacro{} retaining the $K$ elements activated with the largest $|x_i \cdot \|W_{\cdot,i}\|_2 |$\ for\ $i\in\{1,\cdots ,n\}$.
\end{lemma}

\paragraph{Proof.} See Appendix~\ref{app:lemma3.1}.

Building upon our single linear-layer Lemma~\ref{lemma.single_layer}, we now extend it to deep $L$ linear layer networks $\mathcal{M}$ and present that \algacro{} could tighten the gated error upper bound.   

\begin{theorem}[WINA minimizes a provable upper bound on output deviation]
\label{theorem.L_layer_with_activation}
Consider an $L$-layer linear network $\mathcal{M}(\bm{x})=W^{(L)}W^{(L-1)}\cdots W^{(1)}\bm{x}$, where for each $\ell \ge 2$ the weight matrix satisfies $W^{(\ell)\top}W^{(\ell)} = D^{(\ell)}$ with $D^{(\ell)}$ diagonal (column-orthogonality). Let $\mathcal{G}=\{\bm{g}^{(1)},\dots,\bm{g}^{(L)}\}$ be gating variables
with $\bm{g}^{(\ell)}\in\{0,1\}^{d_\ell}$ and
$M^{(\ell)}=\mathrm{diag}(\bm{g}^{(\ell)})$.
Let $\mathcal{M}(\bm{x}\mid\mathcal{G})$ denote the gated network output. Define the output deviation $E(\bm{x};\mathcal{G}):=\|\mathcal{M}(\bm{x})-\mathcal{M}(\bm{x}\mid\mathcal{G})\|_2^2$. Then there exists a separable upper bound $E(\bm{x};\mathcal{G})\le\mathcal{U}(\bm{x};\mathcal{G})$.
Moreover, minimizing $\mathcal{U}(\bm{x};\mathcal{G})$
reduces to selecting, at each layer, the $k$ largest coordinates
weighted by squared column norms. Therefore WINA satisfies
$$
\mathcal{G}_{\mathrm{WINA}}
=
\argmin_{\mathcal{G}}
\mathcal{U}(\bm{x};\mathcal{G}).
$$
\end{theorem}
\paragraph{Proof.} See Appendix~\ref{app:thm3.5} for details.

\paragraph{Remarks on Column-Wise Orthogonality.} Our analysis relies on the column-wise orthogonality of the relevant weight matrices. We applied an efficient one-off offline tensor transformation from~\citep{ashkboos2024slicegptcompresslargelanguage} to enforce column orthogonality. This pre-processing step is lightweight, does not change the functional capacity of the model, and enables our theoretical guarantees to translate effectively to practical settings. We leave details to Appendix \ref{sec:orthogonal_tensor_transform}.


\subsection{Theoretical Validation with Synthetic Experiments}\label{subsec:simulation}

To validate our theoretical analysis, we first conduct controlled synthetic experiments on randomly initialized networks under the assumptions of Lemma~\ref{lemma.single_layer} and Theorem~\ref{theorem.L_layer_with_activation}.

We initialize input vectors and weight matrices using Kaiming initialization with the SiLU activation function, and enforce column-wise orthogonality of weight matrices via the tensor transformation described in Appendix~\ref{sec:orthogonal_tensor_transform}. We then compare the dense network output against sparsified outputs generated by CATS, TEAL, and R-Sparse. Since CATS and TEAL  share the same underlying mechanism and differ only in their sparsity distribution, we group them together as a baseline.

\begin{wraptable}{r}{0.65\linewidth} 
  \vspace{-2.5ex}                      
  \caption{Approximation errors of different methods over randomly initialized networks and sparsity levels. Lower is better.}
  \vspace{-5pt}
  \label{tab:randomly_initialized_matrices}
  \centering
  \small
    \resizebox{0.65\columnwidth}{!}{
    \begin{tabular}{llcccc}
    \toprule[0.1em]
    Theory & Method & 25\% & 40\% & 50\% & 65\% \\ \midrule

    \multirow{3}{*}{Lemma \ref{lemma.single_layer}}
    & CATS/TEAL          & 1.68 ± 0.14 & 3.41 ± 0.20 & 4.86 ± 0.23 & 7.55 ± 0.32  \\
    & R-Sparse           & 1.72 ± 0.13 & 3.48 ± 0.20 & 5.01 ± 0.30 & 7.75 ± 0.35 \\
    & \cellcolor{blue!10} \hspace{-1mm}\textbf{\algacro{}}     & \cellcolor{blue!10}\textbf{0.70 ± 0.05} & \cellcolor{blue!10}\textbf{1.73 ± 0.09} & \cellcolor{blue!10}\textbf{2.70 ± 0.13} & \cellcolor{blue!10}\textbf{4.75 ± 0.15} \\ \midrule

   \multirow{3}{*}{Theorem \ref{theorem.L_layer_with_activation}}
    & CATS/TEAL                    & 0.73 ± 0.04 & 1.44 ± 0.06 & 2.04 ± 0.11 & 3.02 ± 0.18 \\
    & R-Sparse                & 0.77 ± 0.04 & 1.51 ± 0.07 & 2.11 ± 0.11 & 3.13 ± 0.16 \\
    & \cellcolor{blue!10} \hspace{-1mm}\textbf{\algacro{}}     & \cellcolor{blue!10}\textbf{0.38 ± 0.02}  & \cellcolor{blue!10}\textbf{0.76 ± 0.04} & \cellcolor{blue!10}\textbf{1.09 ± 0.06} & \cellcolor{blue!10}\textbf{1.76 ± 0.08} \\ 
  \bottomrule[0.1em]
\end{tabular}
      }
  \vspace{-1ex}
\end{wraptable}
Performance is quantified by the $\ell_2$ deviation between the dense output and the sparsified outputs across varying sparsity ratios. Each experiment is repeated with 20 random seeds. we report averaged results associated with detailed error bar analysis in \autoref{tab:randomly_initialized_matrices}. WINA consistently achieves lower approximation error than competing methods across all sparsity levels and theoretical settings. Notably, WINA reduces error about \textbf{50\%} compared to others, aligning with our theoretical guarantees. Building on this, we turn next to evaluating WINA on LLMs in more realistic settings.

\section{Experiments}
\label{sec:experiments}

\subsection{Setup}
\label{sec:experimental_setup} 
\paragraph{Models.} To demonstrate \algacro{}'s effectiveness across different model families and sizes, we provide our results on four commonly used LLMs: Llama-2-7B~\citep{touvron2023llama}, Llama-3-8B~\citep{dubey2024llama}, Mistral-7B~\citep{jiang2023mistral}, and Phi-4-14B~\citep{abdin2024phi}.

\paragraph{Evaluation.} 
We use the lm-evaluation-harness pipeline~\citep{eval-harness} to assess WINA across a diverse suite of tasks. In addition to the commonsense reasoning examined in works like R-Sparse~\citep{zhang2025r}, we also consider general reasoning, mathematics, and code generation. For commonsense reasoning, we evaluate on PIQA~\citep{bisk2020piqa}, WinoGrande~\citep{sakaguchi2019winogrande}, HellaSwag~\citep{zellers2019hellaswag}, BoolQ~\citep{clark2019boolq}, Arc Challenge and Arc Easy~\citep{allenai:arc}, SciQ~\citep{SciQ}, and OpenBookQA~\citep{OpenBookQA2018}, which measure capabilities such as applying everyday knowledge and resolving coreferences to and answer questions. For general reasoning and knowledge, we consider MMLU~\citep{hendrycks2020measuring}, which spans domains across STEM, humanities, and social sciences, providing a comprehensive test of broad reasoning and knowledge transfer. For math, we adopt GSM8K~\citep{cobbe2021training}, which examines multi-step arithmetic problem solving. For code generation, we use HumanEval~\citep{chen2021evaluating} which requires synthesizing executable Python code. We note that \algacro{} has a closed-form mechanism characterized in Eqn.~(\ref{eq:wina_gate}), making it deterministic given an input. Therefore, there are no error bars to report for WINA's performance.

\paragraph{Baselines.}  
We compare \algacro{} with three recent, representative sparse activation methods: 
CATS~\citep{lee2024catscontextuallyawarethresholdingsparsity}, R-Sparse~\citep{zhang2025r}, and TEAL~\citep{liu2024trainingfreeactivationsparsitylarge}. 
We assign layer-specific sparsity ratios instead of a uniform sparsity across the model so that, given a global sparsity target, we leverage the greedy algorithm as proposed in TEAL~\citep{liu2024trainingfreeactivationsparsitylarge} to iteratively configure per-layer sparsity levels so that the aggregate sparsity meets the global budget. 

\subsection{Results on Commonsense Reasoning} \label{sec:commonsense_reasoning_tasks}
\begin{figure*}[ht!]
  \centering
    \includegraphics[width=\linewidth]{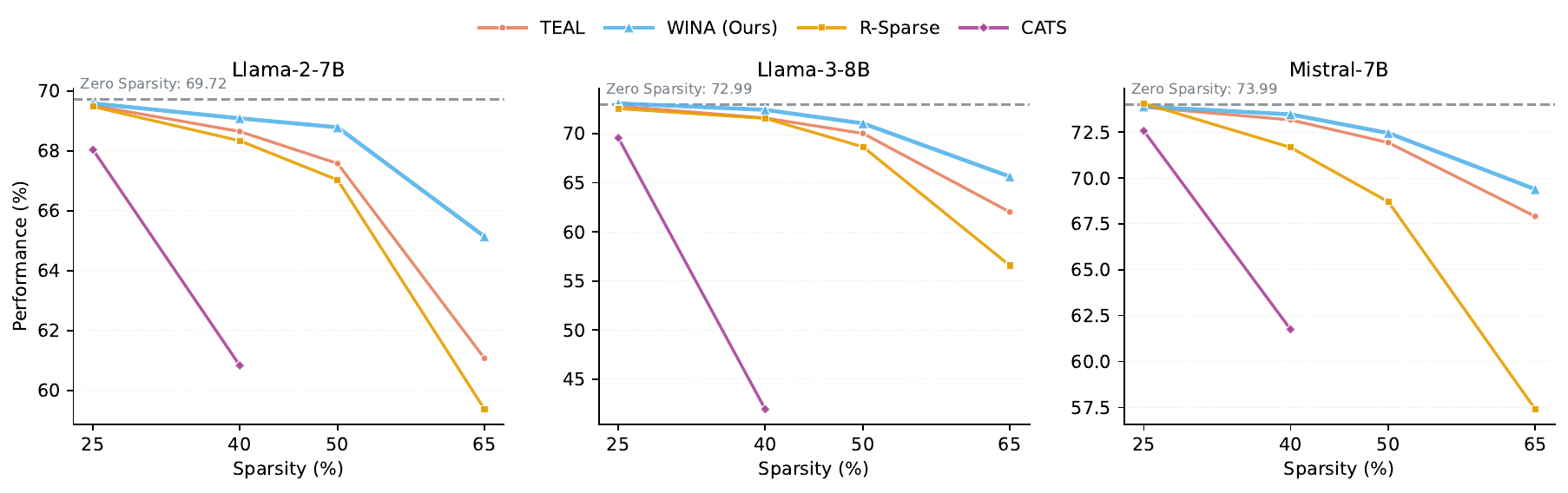}
  \caption{\textbf{Performance-Sparsity Frontier.} Mapping out average performance at each sparsity level for various techniques. \textbf{\textcolor{tabskyblue}{WINA}} outperforms other methods (\textbf{\textcolor{tabcoral}{TEAL}}, \textbf{\textcolor{cAmber}{R-Sparse}}, \textbf{\textcolor{cMagenta}{CATS}}) across sparsity levels with the performance gap increasing as sparsity increases. We note that \textbf{\textcolor{cMagenta}{CATS}} is unable to achieve higher model sparsity due to its applicability to only certain layers.}
  \label{fig:avgperf_sparsity}
\end{figure*}

Following the experimental setup of works like R-Sparse~\citep{zhang2025r}, we provide an empirical comparison of \algacro{} against various baselines (e.g., CATS, R-Sparse, and TEAL) across different sparsity levels, ranging from 25\% to 65\% on commonsense reasoning tasks to demonstrate effectiveness across various experimental settings. \autoref{fig:avgperf_sparsity} provides an comparative overview of the performance-sparsity trade-offs across the different methods.

\begin{table*}[ht]
  \caption{Results over Llama-2-7B on commensense reasoning.}
  \vspace{-5pt}
  \label{tab:llama2_7b_sparsity}
  \centering
    \resizebox{1.0\columnwidth}{!}{
    \begin{tabular}{llccccccccccc}
    \toprule[0.1em]
    Sparsity & Method & PiQA & Arc-C & WinoGrande & HellaSwag & SciQ & OBQA & BoolQ & Arc-E & Avg\\ \midrule

    0\% & Baseline (full model)     & 79.05 & 46.25 & 68.90 & 76.00 & 91.00 & 44.20 & 77.77 & 74.58 & 69.72\\ \midrule

    \multirow{3}{*}{25\%\,}
    & CATS$^\dagger$          & 77.20 & 44.20 & 67.48 & 75.75 & 90.10 & 43.40 & 74.31 & 71.89 & 68.04\\
    & R-Sparse & 78.89 & 45.99 & 68.35 & 75.60 & 91.70 & 43.00 & 77.71 & 74.58 & 69.48\\
    & TEAL                    & 78.73 & 45.99 & 68.90 & 75.98 & 91.00 & 44.20 & 77.09 & 74.24 & 69.52\\
    & \cellcolor{blue!10} \hspace{-1mm}\textbf{\algacro{}}     & \cellcolor{blue!10} \hspace{-1mm}78.40 
    & \cellcolor{blue!10} \hspace{-1mm}46.16 
    & \cellcolor{blue!10} \hspace{-1mm}69.38 
    & \cellcolor{blue!10} \hspace{-1mm}75.93 
    & \cellcolor{blue!10} \hspace{-1mm}90.90 
    & \cellcolor{blue!10} \hspace{-1mm}44.00 
    & \cellcolor{blue!10} \hspace{-1mm}77.03 
    & \cellcolor{blue!10} \hspace{-1mm}74.92 
    & \cellcolor{blue!10} \hspace{-1mm}\textbf{69.59}\\ \midrule

    \multirow{4}{*}{40\%\,}
    & CATS$^\dagger$                    & 74.70 & 36.86 & 61.88 & 69.25 & 79.90 & 39.00 & 66.06 & 58.96 & 60.83\\
    & R-Sparse                & 78.18 & 43.60 & 67.09 & 74.57 & 91.40 & 42.80 & 75.72 & 73.32 & 68.34 \\
    & TEAL                    & 78.07 & 45.22 & 66.85 & 75.31 & 91.40 & 43.20 & 76.48 & 72.69 & 68.65\\
    & \cellcolor{blue!10} \hspace{-1mm}\textbf{\algacro{}} 
    & \cellcolor{blue!10} \hspace{-1mm}78.40 
    & \cellcolor{blue!10} \hspace{-1mm}45.31 
    & \cellcolor{blue!10} \hspace{-1mm}68.59 
    & \cellcolor{blue!10} \hspace{-1mm}75.48 
    & \cellcolor{blue!10} \hspace{-1mm}91.50 
    & \cellcolor{blue!10} \hspace{-1mm}42.40 
    & \cellcolor{blue!10} \hspace{-1mm}76.79 
    & \cellcolor{blue!10} \hspace{-1mm}74.28 
    & \cellcolor{blue!10} \hspace{-1mm}\textbf{69.09}\\ \midrule

    \multirow{3}{*}{50\%\,}
    & R-Sparse                & 76.93 & 42.32 & 65.82 & 72.25 & 91.60 & 41.00 & 74.46 & 71.84 & 67.03 \\
    & TEAL                    & 77.09 & 42.92 & 68.03 & 73.54 & 90.20 & 43.20 & 74.07 & 71.55 & 67.58\\
    & \cellcolor{blue!10} \hspace{-1mm}\textbf{\algacro{}}
    & \cellcolor{blue!10} \hspace{-1mm}77.80 
    & \cellcolor{blue!10} \hspace{-1mm}44.71 
    & \cellcolor{blue!10} \hspace{-1mm}68.51 
    & \cellcolor{blue!10} \hspace{-1mm}74.43 
    & \cellcolor{blue!10} \hspace{-1mm}91.00 
    & \cellcolor{blue!10} \hspace{-1mm}44.60 
    & \cellcolor{blue!10} \hspace{-1mm}75.84 
    & \cellcolor{blue!10} \hspace{-1mm}73.44 
    & \cellcolor{blue!10} \hspace{-1mm}\textbf{68.79}\\ \midrule

   \multirow{3}{*}{65\%\,}
    & R-Sparse & 71.93 & 34.22 & 59.04 & 59.63 & 87.50 & 34.20 & 66.91 & 61.53 & 59.37\\
    & TEAL                    & 73.61 & 36.09 & 62.12 & 63.58 & 84.00 & 38.00 & 69.24 & 61.95 & 61.07\\
    & \cellcolor{blue!10} \hspace{-1mm}\textbf{\algacro{}}     & \cellcolor{blue!10} \hspace{-1mm}75.35 
    & \cellcolor{blue!10} \hspace{-1mm}39.68 
    & \cellcolor{blue!10} \hspace{-1mm}65.82 
    & \cellcolor{blue!10} \hspace{-1mm}69.03 
    & \cellcolor{blue!10} \hspace{-1mm}89.50 
    & \cellcolor{blue!10} \hspace{-1mm}40.40 
    & \cellcolor{blue!10} \hspace{-1mm}72.78 
    & \cellcolor{blue!10} \hspace{-1mm}68.60 
    & \cellcolor{blue!10} \hspace{-1mm}\textbf{65.14}\\
  \bottomrule[0.1em]
\end{tabular}
      }
      \\[0.3em]
    {\fontsize{8}{10}\selectfont
    \begin{tabular}{@{}p{0.9\textwidth}@{}}
    $\dagger$ CATS is unable to reach 50\% or 65\% sparsity since it only achieves sparse activations over MLP layers. \\
    \end{tabular}
    }
\end{table*}

\paragraph{Llama-2-7B.} {On Llama-2-7B, \algacro{} shows strong performance under various sparsity constraints. As shown in \autoref{tab:llama2_7b_sparsity}, \algacro{} achieves the highest average accuracy at 25\% sparsity, outperforming  CATS, R-Sparse and TEAL. While performance naturally degrades at the more extreme 65\% sparsity level, \algacro{} still offers the best accuracy, scoring \textbf{\textcolor{goodgreen}{+5.77\%}} higher than R-Sparse and \textbf{\textcolor{goodgreen}{+4.07\%}} higher than TEAL, suggesting its robustness under aggressive pruning.}

\begin{table*}
  \caption{Results over Llama-3-8B on commonsense reasoning.}
  \vspace{-5pt}
  \label{tab:llama3_8b_sparsity}
  \centering
\resizebox{1.0\columnwidth}{!}{
  \begin{tabular}{llccccccccccc}
    \toprule[0.1em]
    Sparsity & Method & PiQA & Arc-C & WinoGrande & HellaSwag & SciQ & OBQA & BoolQ & Arc-E & Avg \\
    \midrule
    0\% & Baseline (full model) & 80.79 & 53.33 & 72.61 & 79.17 & 93.90 & 45.00 & 81.38 & 77.74 & 72.99\\ \midrule

    \multirow{4}{*}{25\%\,}
    & CATS$^\dagger$          & 78.62 & 48.04 & 70.64 & 76.32 & 91.90 & 41.80 & 78.13 & 71.09 & 69.57\\
    & R-Sparse & 79.82 & 52.05 & 72.38 & 78.69 & 93.50 & 44.40 & 80.92 & 78.75 & 72.56\\
    & TEAL                    & 80.20 & 53.16 & 73.32 & 78.85 & 94.10 & 45.20 & 80.83 & 76.89 & 72.82 \\
    & \cellcolor{blue!10} \hspace{-1mm}\textbf{\algacro{}}     
    & \cellcolor{blue!10} \hspace{-1mm}80.41 
    & \cellcolor{blue!10} \hspace{-1mm}52.82 
    & \cellcolor{blue!10} \hspace{-1mm}73.80 
    & \cellcolor{blue!10} \hspace{-1mm}78.99 
    & \cellcolor{blue!10} \hspace{-1mm}94.00 
    & \cellcolor{blue!10} \hspace{-1mm}44.60 
    & \cellcolor{blue!10} \hspace{-1mm}82.05 
    & \cellcolor{blue!10} \hspace{-1mm}78.03 
    & \cellcolor{blue!10} \hspace{-1mm}\textbf{73.09}\\ \midrule

    \multirow{4}{*}{40\%\,}
    & CATS$^\dagger$                    & 59.96 & 27.82 & 51.30 & 40.18 & 46.10 & 29.80 & 42.26 & 38.09 & 41.94 \\
    & R-Sparse                & 79.05 & 50.26 & 72.14 & 76.91 & 94.10 & 43.00 & 79.14 & 77.86 & 71.56\\
    & TEAL                    & 79.00 & 48.98 & 71.82 & 77.45 & 93.30 & 45.00 & 80.03 & 77.19 & 71.60\\
    & \cellcolor{blue!10} \hspace{-1mm}\textbf{\algacro{}}     
    & \cellcolor{blue!10} \hspace{-1mm}79.87 
    & \cellcolor{blue!10} \hspace{-1mm}50.68 
    & \cellcolor{blue!10} \hspace{-1mm}72.30 
    & \cellcolor{blue!10} \hspace{-1mm}77.91 
    & \cellcolor{blue!10} \hspace{-1mm}93.90 
    & \cellcolor{blue!10} \hspace{-1mm}45.00 
    & \cellcolor{blue!10} \hspace{-1mm}82.23 
    & \cellcolor{blue!10} \hspace{-1mm}77.57 
    & \cellcolor{blue!10} \hspace{-1mm}\textbf{72.43}\\ \midrule

    \multirow{3}{*}{50\%\,}
    & R-Sparse                & 76.22 & 45.73 & 66.61 & 73.22 & 93.80 & 42.20 & 76.70 & 74.83  & 68.66\\
    & TEAL                    & 78.29 & 48.12 & 70.09 & 74.83 & 93.70 & 42.60 & 78.23 & 74.41 & 70.03\\
    & \cellcolor{blue!10} \hspace{-1mm}\textbf{\algacro{}}     
    &  \cellcolor{blue!10} \hspace{-1mm}79.16 
    & \cellcolor{blue!10} \hspace{-1mm}48.81 
    & \cellcolor{blue!10} \hspace{-1mm}70.64 
    & \cellcolor{blue!10} \hspace{-1mm}76.44 
    & \cellcolor{blue!10} \hspace{-1mm}93.50 
    & \cellcolor{blue!10} \hspace{-1mm}43.60 
    & \cellcolor{blue!10} \hspace{-1mm}81.25 
    & \cellcolor{blue!10} \hspace{-1mm}75.00 
    & \cellcolor{blue!10} \hspace{-1mm}\textbf{71.05}\\ \midrule

   \multirow{3}{*}{65\%\,}
   & R-Sparse & 68.50 & 33.36 & 57.38 & 51.48 & 86.00 & 31.80 & 65.23 & 58.80 & 56.57\\
    & TEAL & 73.34 & 37.37 & 63.46 & 61.76 & 88.90 & 37.00 & 69.85 & 64.48 & 62.02 \\    
    & \cellcolor{blue!10} \hspace{-1mm}\textbf{\algacro{}}     & \cellcolor{blue!10} \hspace{-1mm}74.65 & \cellcolor{blue!10} \hspace{-1mm}41.98 
    & \cellcolor{blue!10} \hspace{-1mm}64.48 
    & \cellcolor{blue!10} \hspace{-1mm}67.89 
    & \cellcolor{blue!10} \hspace{-1mm}90.70 
    & \cellcolor{blue!10} \hspace{-1mm}41.60 
    & \cellcolor{blue!10} \hspace{-1mm}76.73 
    & \cellcolor{blue!10} \hspace{-1mm}67.00 
    & \cellcolor{blue!10} \hspace{-1mm}\textbf{65.63}\\

    \bottomrule[0.1em]
  \end{tabular}
    }
    \\[0.3em]
    {\fontsize{8}{10}\selectfont
    \begin{tabular}{@{}p{0.9\textwidth}@{}}
    $\dagger$ CATS is unable to reach 50\% or 65\% sparsity since it only achieves sparse activations over MLP layers. \\
    \end{tabular}
    }
\end{table*}
\paragraph{Llama-3-8B.} 
Across all sparsity levels, \algacro{} (\autoref{tab:llama3_8b_sparsity}) consistently outperforms or matches the best baseline methods. At 25\% sparsity, \algacro{} achieves the highest average score of 73.09\%, surpassing the full baseline model (72.99\%) and other sparse methods. This trend continues at higher sparsity levels: \algacro{} maintains strong performance at 40\% and 50\% sparsity with average scores of 72.43\% and 71.05\%, respectively. Even at 65\% sparsity, \algacro{} remains competitive while other methods experience significant degradation, on average scoring \textbf{\textcolor{goodgreen}{+3.61\%}} higher than TEAL and \textbf{\textcolor{goodgreen}{+9.06\%}} higher than R-Sparse.

\begin{table*}[ht]
  \caption{Results over Mistral-7B on commensense reasoning.}
  \vspace{-5pt}
  \label{tab:mistral7b_sparsity}
  \centering
    \resizebox{1.0\columnwidth}{!}{
    \begin{tabular}{llccccccccccc}
    \toprule[0.1em]
    Sparsity & Method & PiQA & Arc-C & WinoGrande & HellaSwag & SciQ & OBQA & BoolQ & Arc-E & Avg\\ \midrule

    0\% & Basline (full model)     & 82.05 & 54.01 & 73.88 & 81.06 & 93.90 & 43.80 & 83.61 & 79.59 & 73.99
\\ \midrule

    \multirow{4}{*}{25\%\,}
    & CATS$^\dagger$ & 81.01 & 52.22 & 72.38 & 80.86 & 92.00 & 42.40 & 82.60 & 77.06  & 72.57\\
    & R-Sparse & 81.61 & 53.41 & 74.19 & 81.06 & 94.10 & 45.60 & 83.73 & 78.83 & \textbf{74.07} \\
    & TEAL                    & 82.21 & 53.75 & 73.48 & 80.93 & 94.10 & 44.00 & 83.52 & 78.87 & 73.86\\
    & \cellcolor{blue!10} \hspace{-1mm}\textbf{\algacro{}}     & \cellcolor{blue!10}\hspace{-1mm}82.10 
    & \cellcolor{blue!10} \hspace{-1mm}53.24 
    & \cellcolor{blue!10} \hspace{-1mm}74.35 
    & \cellcolor{blue!10} \hspace{-1mm}80.88 
    & \cellcolor{blue!10} \hspace{-1mm}93.90 
    & \cellcolor{blue!10} \hspace{-1mm}43.60 
    & \cellcolor{blue!10} \hspace{-1mm}83.30 
    & \cellcolor{blue!10} \hspace{-1mm}79.80 
    & \cellcolor{blue!10} \hspace{-1mm}73.90\\ \midrule

    \multirow{4}{*}{40\%\,}
    & CATS$^\dagger$                     &  75.95 & 41.13 & 63.61 & 69.27 & 78.90 & 36.20 & 68.07 & 60.86 & 61.75\\
    & R-Sparse   &  81.77 & 51.62 & 70.09 & 77.56 & 91.50 & 39.20 & 83.09 & 78.54 & 71.67 \\
    & TEAL                    & 81.56 & 52.30 & 73.40 & 80.03 & 93.80 & 43.00 & 83.21 & 78.07 & 73.17\\
    & \cellcolor{blue!10} \hspace{-1mm}\textbf{\algacro{}}     & \cellcolor{blue!10} \hspace{-1mm}82.05 & \cellcolor{blue!10} \hspace{-1mm}53.33 
    & \cellcolor{blue!10} \hspace{-1mm}72.38 
    & \cellcolor{blue!10} \hspace{-1mm}80.41 
    & \cellcolor{blue!10} \hspace{-1mm}94.30 
    & \cellcolor{blue!10} \hspace{-1mm}43.60 
    & \cellcolor{blue!10} \hspace{-1mm}82.87 
    & \cellcolor{blue!10} \hspace{-1mm}78.83 
    & \cellcolor{blue!10} \hspace{-1mm}\textbf{73.47}\\ \midrule

    \multirow{3}{*}{50\%\,}
    & R-Sparse                & 81.07 & 49.74 & 66.22 & 69.39 & 89.50 & 34.20 & 82.14 & 77.36 & 68.70 \\
    & TEAL                    & 79.76 & 49.57 & 70.80 & 78.84 & 93.90 & 43.00 & 82.75 & 76.81 & 71.93\\
    & \cellcolor{blue!10} \hspace{-1mm}\textbf{\algacro{}}     & \cellcolor{blue!10} \hspace{-1mm}81.34 
    & \cellcolor{blue!10} \hspace{-1mm}52.30 
    & \cellcolor{blue!10} \hspace{-1mm}70.88 
    & \cellcolor{blue!10} \hspace{-1mm}79.60 
    & \cellcolor{blue!10} \hspace{-1mm}94.50 
    & \cellcolor{blue!10} \hspace{-1mm}41.80 
    & \cellcolor{blue!10} \hspace{-1mm}81.59 
    & \cellcolor{blue!10} \hspace{-1mm}77.57 
    & \cellcolor{blue!10} \hspace{-1mm}\textbf{72.45}\\ \midrule

   \multirow{3}{*}{65\%\,}
    & R-Sparse & 63.60 & 36.52 & 63.77 & 59.30 & 88.00 & 30.20 & 74.16 & 43.64 & 57.40\\
    & TEAL                    & 77.97 & 42.49 & 66.22 & 72.26 & 92.20 & 40.40 & 79.27 & 72.47 & 67.91 \\
    & \cellcolor{blue!10} \hspace{-1mm}\textbf{\algacro{}}     
    & \cellcolor{blue!10} \hspace{-1mm}78.24 
    & \cellcolor{blue!10} \hspace{-1mm}48.12 
    & \cellcolor{blue!10} \hspace{-1mm}66.38 
    & \cellcolor{blue!10} \hspace{-1mm}75.11 
    & \cellcolor{blue!10} \hspace{-1mm}92.90 
    & \cellcolor{blue!10} \hspace{-1mm}42.00 
    & \cellcolor{blue!10} \hspace{-1mm}77.65 
    & \cellcolor{blue!10} \hspace{-1mm}74.75 
    & \cellcolor{blue!10} \hspace{-1mm}\textbf{69.39}\\
  \bottomrule[0.1em]
\end{tabular}
      }
      \\[0.3em]
    {\fontsize{8}{10}\selectfont
    \begin{tabular}{@{}p{0.9\textwidth}@{}}
    $\dagger$ CATS is unable to reach 50\% or 65\% sparsity since it only achieves sparse activations over MLP layers. \\
    \end{tabular}
    }
\end{table*}

\paragraph{Mistral-7B.} Overall, \algacro{} matches or outperforms or matches the best performing methods across different sparsity levels (\autoref{tab:mistral7b_sparsity}). At moderate sparsity (25-40\%), \algacro{} maintains performance nearly identical to the full model baseline, showing strong robustness to pruning. \algacro{}'s performance becomes most evident under extreme sparsity: at 65\% sparsity, \algacro{} achieves an average score of 69.39\%, which is \textbf{\textcolor{goodgreen}{+9.48\%}} higher than R-Sparse (59.91\%) and \textbf{\textcolor{goodgreen}{+1.48\%}} higher than TEAL (67.91\%), demonstrating that \algacro{} is more effective under aggressive compression.

\subsection{Additional Analysis And Ablations} 

\paragraph{Additional Benchmarks and Architecture.} We conduct additional experiments on several advanced tasks, including general reasoning (MMLU), mathematics (GSM8K), and coding (HumanEval). Additionally, we use Phi-4-14B to test the adaptability of \algacro{} to different model architectures. To focus our analysis and reduce clutter, we compare against TEAL, the second best performing method in Section~\ref{sec:commonsense_reasoning_tasks} as an upper-bound. We detail our results in \autoref{tab:phi_4_sparsity}. 

We observe that \algacro{} consistently delivers better performance than TEAL across all sparsity levels and tasks, highlighting its robustness on diverse applications, ranging from commonsense reasoning and subject knowledge to mathematical reasoning and code generation. Similar to our findings from Section \ref{sec:commonsense_reasoning_tasks}, \algacro{}'s performance advantage becomes even more pronounced at higher sparsity levels, demonstrating its superior scalability. At 65\% sparsity, the improvements over TEAL are considerable: \textbf{\textcolor{goodgreen}{+1.65\%}} for Commonsense Reasoning, \textbf{\textcolor{goodgreen}{+4.88\%}} for MMLU, \textbf{\textcolor{goodgreen}{+2.73\%}} for GSM8K, and  \textbf{\textcolor{goodgreen}{+9.14\%}} for HumanEval. Notably, on HumanEval, \algacro{} even surpasses the baseline, a phenomenon often observed in pruning literature (e.g., see Section~\ref{sec:related_work}) as its sparse activation may also be implicitly regularizing or suppressing harmful neurons, thereby enhancing performance.

\begin{table*}[ht]
  \caption{Results over Phi-4-14B over additional benchmarks.}
  \vspace{-7pt}
  \label{tab:phi_4_sparsity}
  \centering
  \scriptsize
    \resizebox{0.75\columnwidth}{!}{
    \begin{tabular}{llcccccc}
    \toprule[0.1em]
    Sparsity & Method & Commonsense (avg) & MMLU & GSM8K & HumanEval\\ \midrule

    0\% & Baseline (full model)     & 74.38 & 77.06 & 90.22 & 50.61
\\ \midrule

    \multirow{2}{*}{25\%\,}
    & TEAL                    & 74.12 & 76.63 & 89.84 & 46.95\\
    & \cellcolor{blue!10} \hspace{-1mm}\textbf{\algacro{}}     
    & \cellcolor{blue!10} \hspace{-1mm}\textbf{74.25} & \cellcolor{blue!10} \hspace{-2mm} \textbf{76.60} & \cellcolor{blue!10} \hspace{-2mm} \textbf{90.22}  & \cellcolor{blue!10} \hspace{-2mm} \textbf{50.00}\\ \midrule

    \multirow{2}{*}{40\%\,}
    & TEAL                    & 73.49 & 75.10 & 88.02 & 45.73\\
    & \cellcolor{blue!10} \hspace{-1mm}\textbf{\algacro{}}    
    & \cellcolor{blue!10} \hspace{-1mm}\textbf{73.76}  & \cellcolor{blue!10} \hspace{-2mm} \textbf{76.44}  & \cellcolor{blue!10} \hspace{-2mm} \textbf{90.67}  & \cellcolor{blue!10} \hspace{-2mm} \textbf{53.00}\\ \midrule

    \multirow{2}{*}{50\%\,}
    & TEAL                    &  72.73 & 73.52 & 86.13 & 41.46\\
    & \cellcolor{blue!10} \hspace{-1mm}\textbf{\algacro{}}     
    
    & \cellcolor{blue!10} \hspace{-1mm}\textbf{73.51}  & \cellcolor{blue!10} \hspace{-2mm} \textbf{75.83}  & \cellcolor{blue!10} \hspace{-2mm} \textbf{87.57}  & \cellcolor{blue!10} \hspace{-2mm} \textbf{51.83}\\ \midrule

   \multirow{2}{*}{65\%\,}
    & TEAL                    &   69.93 & 65.17 & 74.37 & 32.32\\
    & \cellcolor{blue!10} \hspace{-1mm}\textbf{\algacro{}}     
    & \cellcolor{blue!10} \hspace{-1mm}\textbf{71.58}  & \cellcolor{blue!10} \hspace{-2mm} \textbf{70.05}  & \cellcolor{blue!10} \hspace{-2mm} \textbf{77.10}  & \cellcolor{blue!10} \hspace{-2mm} \textbf{41.46}\\
  \bottomrule[0.1em]
\end{tabular}
\vspace{-3mm}
      }
\end{table*}

\begin{figure}[ht!]
    \centering
        \includegraphics[width=0.85\linewidth]{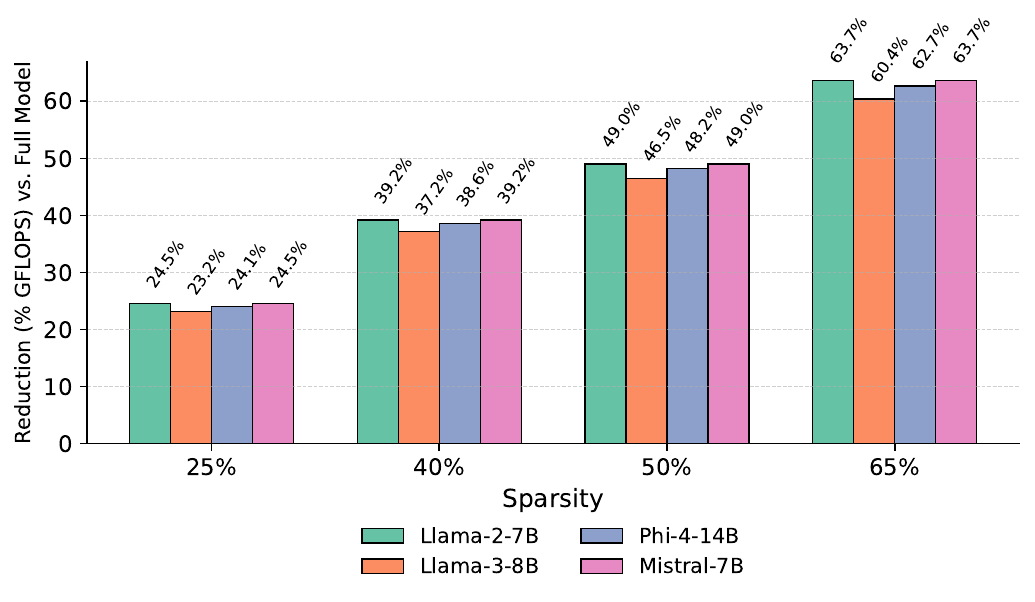}
    \caption{\textbf{Computational Savings from WINA.} Percentages indicate the reduction in GFLOPs at the specified sparsity level vs. the full dense model (higher is better). 
    }
    \label{fig:gflops}
    \vspace{-4mm}
\end{figure}

\paragraph{Computational Savings.} In addition to performance gains, \algacro{} also yields computational savings; as shown in~\autoref{fig:gflops}, \algacro{} reduces the overall (G)FLOPs by up to 63.7\% on Llama-2-7B and Mistral-7B, 60.4\% on Llama-3-8B, and 62.7\% on Phi-4-14B at 65\% sparsity, potentially translating to faster inference and lower computational costs under tight resource constraints. 

\begin{wraptable}{r}{0.5\linewidth}
  \caption{{ Llama-2-7B results across sparsity and quantization levels on commonsense reasoning.}}
  \label{tab:quantized_model}
  \centering
  \resizebox{0.45\columnwidth}{!}{
    \begin{tabular}{llccc}
      \toprule[0.1em]
      Sparsity & Method & $\text{FP}_{16}$ & $\text{INT}_{8}$ & $\text{INT}_{4}$ \\
      \midrule
       0\% & Baseline (full model) & 69.72 & 69.39 & 68.06 \\\midrule

      \multirow{2}{*}{25\%} & TEAL & 69.52 & 68.09 & 67.93 \\
       & \cellcolor{blue!10}\textbf{\algacro{}} & \cellcolor{blue!10}\textbf{69.59} & \cellcolor{blue!10}\textbf{68.39} & \cellcolor{blue!10}\textbf{68.06} \\\midrule

      \multirow{2}{*}{40\%} & TEAL & 68.65 & 67.66 & 67.39 \\
       & \cellcolor{blue!10}\textbf{\algacro{}} & \cellcolor{blue!10}\textbf{69.09} & \cellcolor{blue!10}\textbf{68.16} & \cellcolor{blue!10}\textbf{67.52} \\\midrule

      \multirow{2}{*}{50\%} & TEAL & 67.58 & 66.41 & 66.00 \\
       & \cellcolor{blue!10}\textbf{\algacro{}} & \cellcolor{blue!10}\textbf{68.79} & \cellcolor{blue!10}\textbf{67.28} & \cellcolor{blue!10}\textbf{66.50} \\\midrule

      \multirow{2}{*}{65\%} & TEAL & 61.07 & 60.38 & 59.99 \\
       & \cellcolor{blue!10}\textbf{\algacro{}} & \cellcolor{blue!10}\textbf{65.14} & \cellcolor{blue!10}\textbf{64.24} & \cellcolor{blue!10}\textbf{63.67} \\
      \bottomrule[0.1em]
      \vspace{-5mm}
    \end{tabular}
  }
\end{wraptable}
\paragraph{Compatibility with Quantization.}  In resource constrained environments, LLMs are typically deployed in quantized form or lower precision~\citep{chen2023efficientllm}. Although prior works rarely evaluate the robustness of their methods under quantization, we show that \algacro{} is naturally compatible and apply it onto quantized versions of Llama-2-7B to assess performance on diverse commonsense reasoning tasks. 
As shown in \autoref{tab:quantized_model}, \algacro{} incurs only minimal degradation, consistently outperforming TEAL; similar to before, the performance gap between WINA and TEAL increases as sparsity increases. Most notably, at 65\% sparsity, WINA outperforms TEAL by 
\textbf{\textcolor{goodgreen}{+4.07\%}} on 16-bit, \textbf{\textcolor{goodgreen}{+3.86\%}} on 8-bit, and \textbf{\textcolor{goodgreen}{+3.68\%}} on 4-bit quantized models. We leave the treatment of sparsification-aware quantization (e.g., joint sparse-activation-quantization~\citep{lin2024awq}), to future work.

\begin{wraptable}{r}{0.5\linewidth}
  \vspace{-5mm}
  \caption{Ablating the effects of orthogonalization.}
  \label{tab:ablation_study}
  \centering
    \resizebox{0.45\columnwidth}{!}{
    \begin{tabular}{llcc}
  \toprule[0.1em]
  Method & Sparsity & Llama-2-7B & Llama-3-8B \\
  \midrule
  \multirow{2}{*}{25\%} 
  & TEAL (transform) & \textbf{69.70} & 72.91   \\
  & \cellcolor{blue!10}\textbf{\algacro{}} & \cellcolor{blue!10}69.59  & \cellcolor{blue!10}\textbf{73.09}   \\\midrule
  \multirow{2}{*}{40\%} 
  & TEAL (transform) & \textbf{69.11}  & 72.05   \\
  & \cellcolor{blue!10}\textbf{\algacro{}} & \cellcolor{blue!10}69.09 & \cellcolor{blue!10}\textbf{72.43}   \\\midrule
  \multirow{2}{*}{50\%} 
  & TEAL (transform) & 67.97  & 70.36   \\
  & \cellcolor{blue!10}\textbf{\algacro{}} & \cellcolor{blue!10}\textbf{68.79}  & \cellcolor{blue!10}\textbf{71.05}   \\\midrule
  \multirow{2}{*}{65\%} 
  & TEAL (transform) & 63.05  &  63.52  \\
  & \cellcolor{blue!10}\textbf{\algacro{}} & \cellcolor{blue!10}\textbf{65.14}  &  \cellcolor{blue!10}\textbf{65.63}  \\  
  \bottomrule[0.1em]
\end{tabular}
      }
\end{wraptable}
\paragraph{Ablative Study on Orthogonalization.} To control for potential confounding effects from the transformation process, we introduce an additional baseline, TEAL-Transform, where TEAL is also applied to the transformed model, retaining the $K$ elements with the largest absolute values $|\bm{x}|$.  \autoref{tab:ablation_study} shows the average performance over commonsense datasets from our main experiments. 
Consistent with earlier results, \algacro{} still outperforms in most cases with the gap widening as sparsity increases, demonstrating that gains arise primarily from \algacro{}'s gating function.

\paragraph{Social Bias.} To account for the effect of model compression on potential social/model biases, following \citep{strubell2023biascompress}, we evaluate our sparsified models on CrowS-Pairs (\autoref{tab:bias-sparsity}). Across all models and sparsity levels, we observe no systematic increase in bias, suggesting \algacro{} does not exacerbate bias and can modestly mitigate it. We leave full details to Appendix \ref{sec:social_bias}.

\subsection{Hardware Acceleration}\label{sec:hardware_acceleration}

To achieve realistic speed-ups, we develop a dedicated sparse GEMV kernel for \algacro{} using Triton~\citep{tillet2019triton}. This kernel takes the input tensor $\bm{x}$, matrix $W$, pre-computed column-wise norm $\bm{c}$ of $W$, and a desired sparsity level to return an output $\bm{x}\odot\bm{g}$. \algacro{} gating $\bm{g}$ is computed via Eq. (\ref{eq:wina_gate}), selecting a subset of columns of $W$ before the input $\bm{x}$ is fed into the matrix multiply. The weight norms $\bm{c}$ are pre-computed offline once during model loading. Compared to TEAL's sparse kernel, \algacro{} only introduces an additional element-wise product $\bm{x}\odot\bm{c}$, whose cost is negligible. 

In particular, let $d$ be the hidden dimension, $B$ be the batch size, and $T$ be the sequence length so that a standard linear layer incurs $\mathcal{O}(BTd^2)$ FLOPs during the forward pass while \algacro{}'s gating mechanism requires $\mathcal{O}(BTd)$ FLOPs. Comparing the two, their ratio is on order of $\mathcal{O}(BTd)/\mathcal{O}(BTd^2)=\mathcal{O}(1/d)$.  Since $d$ is typically large in modern LLMs, (e.g, 2048, 4096 or larger), \algacro{} consumes only $\ll 0.1\%$ additional overhead, which is negligible compared to the acceleration and accuracy improvements that its sparsification provides.

\begin{figure}[ht]
    \centering
    \includegraphics[width=.95\linewidth]{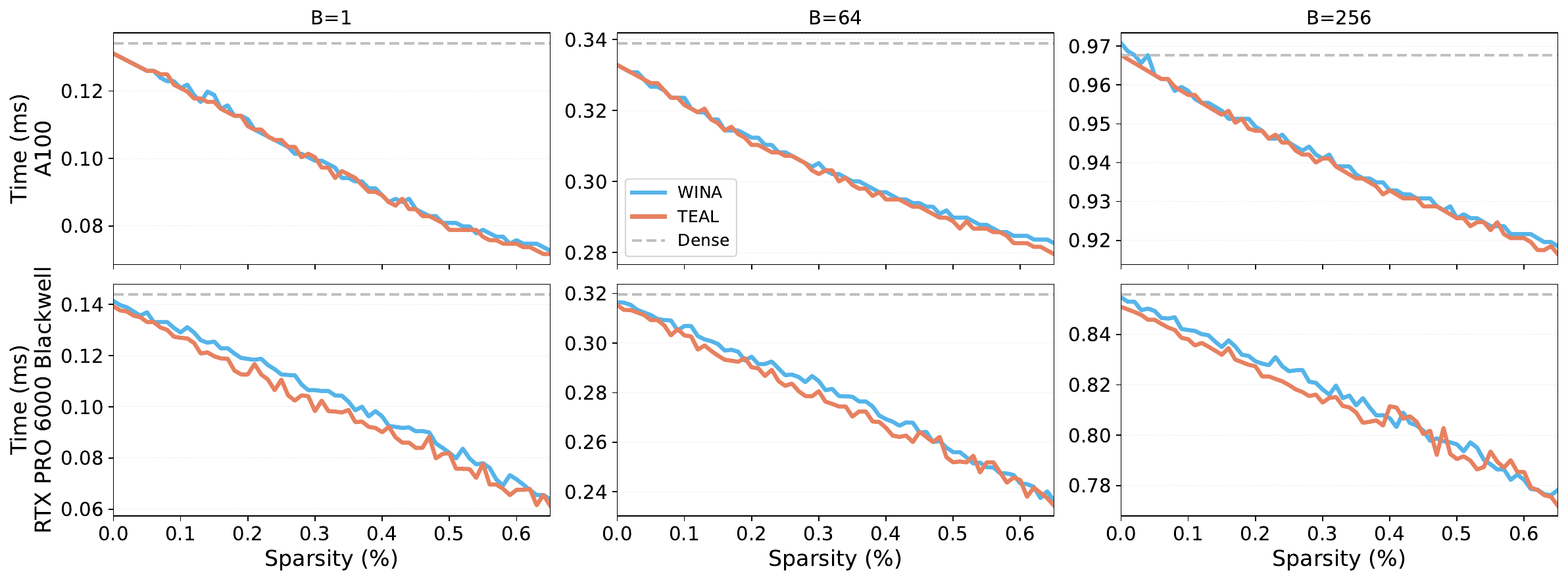}    
    \caption{Sparsity vs. latency across different batch sizes $B \in\{1, 64, 256\}$ for GEMV (generalized matrix-vector multiplication) of sizes $5120\times1$ and $5120\times 17920$. WINA's Triton kernel performance consistently matches that of TEAL across different GPU architectures (A100 top, RTX PRO 6000 Blackwell bottom), achieving similar speedups across our sparsity levels and as sparsity increases.}
    \label{fig:a100_rtx_kernel}
\end{figure}

To empirically verify the negligible runtime overhead, we benchmark our Triton \algacro{} kernel across commonly used matrix shapes in modern LLMs and batch sizes ${1, 64, 256}$. To avoid hardware-specific bias, we profile on three GPU architectures: A800, A100, and the RTX PRO 6000 Blackwell. Results in \autoref{fig:a100_rtx_kernel} show that \algacro{} matches TEAL’s speed/latency almost identically. This phenomenon is consistent across all tested hardware platforms and batch sizes. Due to space considerations, we leave results on A800 and additional details to Appendix~\ref{sec:addition_hardware_acceleration}. Additionally, we investigate how the speed-up ratio varies with the batch size ( Appendix~\ref{sec:speedup_batch_size}). As batch size increases and GPU memory bandwidth becomes saturated, the achievable speed-up gradually decreases. Nevertheless, degradation is mild and \algacro{} still delivers substantial acceleration, achieving approximately up to 42\% speed-up on A100 and up to 57\% RTX PRO 6000 Blackwell across different batch sizes, similar to TEAL.

\subsection{Performance on Long Context Tasks}
We further evaluate the long-context reasoning ability of models under \algacro{} on LongBench~\citep{bai2024longbench}, a multi-task benchmark with realistic long-context scenarios involving code completion, summarization, as well as single and multi-document QA. LongBench contains 4,750 examples with average task length ranging from 5-15K.
We evaluated Llama-2-7B, Llama-3-8B, and Phi-4-14B, which support context windows of 4K, 8K, and 16K tokens, respectively. \autoref{tab:longbench_llama2} shows results for Llama-2-7B at 4K where, across all settings, \algacro{} consistently outperforms TEAL even in more demanding scenarios. We leave additional results to Appendix \ref{subsec: longbench_results}. 

\begin{table}[htbp]
  \caption{Performance on LongBench (Llama-2-7B-4K).}
  \scriptsize
  \label{tab:longbench_llama2}
  \begin{tabularx}{\textwidth}{lCCCCCCCC}
    \toprule
    Sparsity & Method & 
    Code Completion & Few-shot Learning & Summari-zation &
    Multi-Document QA & Single-Document QA & Synthetic Tasks & Overall \\
    \midrule
     0 & Baseline &
    62.28 & 52.09 & 11.63 &
    6.73 & 13.00 & 5.56 &  22.62 \\\midrule
    \multirow{2}{*}{0.25}
    & TEAL  & 62.14 & 51.80 & 11.83 & 7.24 & 12.87 & 5.06 & 22.59 \\
    & \cellcolor{blue!10}\textbf{WINA} & \cellcolor{blue!10}62.54 & \cellcolor{blue!10}52.46 & \cellcolor{blue!10}12.48 & \cellcolor{blue!10}7.07 &
    \cellcolor{blue!10}12.95 & \cellcolor{blue!10}5.26 &  \cellcolor{blue!10}\textbf{22.89} \\\midrule
    \multirow{2}{*}{0.4}
    & TEAL & 60.28 & 52.23 & 12.55 & 8.12 & 13.40  & 4.54 & 22.83 \\
    & \cellcolor{blue!10}\textbf{WINA} & \cellcolor{blue!10}61.52 & \cellcolor{blue!10}51.76 & \cellcolor{blue!10}13.81 & \cellcolor{blue!10}7.63  & \cellcolor{blue!10}13.64 & \cellcolor{blue!10}4.97 & \cellcolor{blue!10}\textbf{23.11} \\\midrule
    \multirow{2}{*}{0.5}
    & TEAL & 58.41 & 50.10 & 12.97 & 8.39 & 12.80  & 4.28 & 22.23 \\
    & \cellcolor{blue!10}\textbf{WINA} & \cellcolor{blue!10}61.15 & \cellcolor{blue!10}52.16 & \cellcolor{blue!10}14.62 & \cellcolor{blue!10}7.41  & \cellcolor{blue!10}12.09 & \cellcolor{blue!10}3.13 & \cellcolor{blue!10}\textbf{22.71}  \\\midrule
    \multirow{2}{*}{0.65}
    & TEAL & 44.20 & 44.15 & 11.16 & 7.09 & 8.17 & 1.62 & 17.88 \\
    & \cellcolor{blue!10}\textbf{WINA} & \cellcolor{blue!10}55.03 & \cellcolor{blue!10}48.93 & \cellcolor{blue!10}8.98 & \cellcolor{blue!10}6.42  & \cellcolor{blue!10}8.60  & \cellcolor{blue!10}2.85  & \cellcolor{blue!10}\textbf{19.54} \\
    \bottomrule
  \end{tabularx}
\end{table}


\section{Conclusion}\label{sec:conclusion}

\algacro{} offers not only a simple yet highly effective training-free sparse-activation for accelerating LLMs but also solid theoretical guarantees on its approximation error. Our results show that it consistently outperforms strong baselines across many benchmarks, while maintaining robustness even under aggressive 4-bit and 8-bit quantization. These results highlight both the practicality and the plug-and-play usability of \algacro{}, making it ideal for sparse-activation-accelerated inference.

\newpage
\section*{Ethics Statement}\label{sec:ethics}
Our work introduces a training-free sparse activation framework designed to improve the efficiency of large language model inference. The primary ethical consideration is ensuring that such efficiency gains are not misused to scale potentially harmful or biased models more widely without proper safeguards. We emphasize that WINA is a general-purpose acceleration method and does not alter the underlying datasets or model outputs. As such, any societal risks or biases present in the base models remain unchanged, and practitioners must remain vigilant about responsible deployment. Additionally, by lowering computational costs, our approach may promote accessibility of research to institutions with limited resources, contributing to broader inclusivity in AI research.

\section*{Reproducibility Statement}\label{sec:repro}

We have taken steps to ensure the reproducibility of our results. All experiments were conducted on publicly available benchmark datasets. We used standard open-source evaluation pipelines such as lm-evaluation-harness for benchmarking. Detailed experimental settings are reported in Section~\ref{sec:experiments}, including models (\eg, Llama-2, Llama-3, Mistral-7B, and Phi-4), sparsity configurations, and computational resources. Comparisons were made against established baselines under controlled sparsity levels, and we provide theoretical analyses and ablation studies to support our claims.
The total run time of our experiments were run using one A100 80GB GPU for several days along with a few hours on a RTX PRO 6000 Blackwell and A800 for minor benchmarking. Our source code is made anonymously available at \href{https://github.com/microsoft/wina}{URL}.

\bibliography{reference}
\bibliographystyle{iclr2026_conference}


\clearpage
\appendix

\section{Appendix}

\subsection{The Use of Large Language Models (LLMs)}\label{sec:use_llm}

The use of LLMs in this paper was restricted to limited editing supports, such as detecting and correcting grammar errors, typos, rephrasing sentences.

\subsection{Orthogonal Tensor Transformation}
\label{sec:orthogonal_tensor_transform}

To enforce orthogonality in DNNs, there are two families of approaches. The first one is to augment an orthogonality regularizer $\norm{W^\top W -I}$ into the objective function~\citep{xie2017all,huang2020controllable}. This line of approach requires some model training. Another family of approaches uses matrix transformation or rotation to enforce orthogonality. In particular, SliceGPT~\citep{ashkboos2024slicegptcompresslargelanguage} propose a transformation to bring orthogonality while preserve the numerical equivalence or computation invariance. Similar transformation is also widely used in efficient-AI literatures~\citep{ashkboos2024quarot,chee2023quip,liu2024spinquant,hu2025ostquant}. In this work, we adopt the same transformation from \citep{ashkboos2024slicegptcompresslargelanguage} to produce computational invariant LLMs with enforced orthogonality to maximize the performance gain of \algacro{}. 

\paragraph{Overview.} Given a weight matrix $W$, we can enforce column-wise orthogonality by multiplying $W$ from the right by an orthogonal matrix $Q$ such that the product $WQ$ has orthogonal columns. Specifically, we perform Singular Value Decomposition (SVD) on $W$, $W = U \Sigma V^\top$,
where $U$ and $V$ are orthogonal matrices, and $\Sigma$ is a diagonal matrix containing the singular values of $W$. To achieve column-orthogonality, we set $Q = V$ and transform $W$ as $\widehat{W} = W V$.
This transformation guarantees that the resulting matrix $\widehat{W}$ satisfies the column-orthogonality:
\begin{equation}
(\widehat{W})^\top \widehat{W} = \Sigma^\top U^\top U \Sigma = \Sigma^2.
\end{equation}
To ensure that the model's final output remains unchanged after this transformation, we compensate for its effects using computational invariance \citep{ashkboos2024slicegptcompresslargelanguage}. In particular, we enforce column-wise orthogonality constraints over the matrices via SVD-based transformation. Detailed pseudoscope of Algorithm is present \ref{alg:tensor_transformation}. Without loss of generality, we present pseudocode to a transformer-based model $\mathcal{M}$ equipping with $L$ layers. Each layer includes the following weight matrices: $\left\{W_k^{(\ell)}, W_q^{(\ell)}, W_v^{(\ell)}, W_o^{(\ell)}, W_{\text{gate}}^{(\ell)}, W_{\text{up}}^{(\ell)}, W_{\text{down}}^{(\ell)}\right\}$ for $\ell = 1, \dots, L$, along with the output projection matrix $W_{\text{head}}$ of the final head layer. 

\paragraph{Runtime.} This transformation is lightweight and efficient, taking less than four minutes on a single A100 GPU for models like Llama-2-3B, Llama-3-8B, and Mistral, and less than twelve minutes for Phi-4-14B.

\begin{algorithm}
\setstretch{1.2}
\caption{Orthogonal Tensor Transformation}
\label{alg:tensor_transformation}
\begin{algorithmic}[1]
\State \textbf{Input:} Model $\mathcal{M}$ with matrix $W_{emb}$ of embedding layer, $L$ layers with matrices $\{W_k^{(\ell)}, W_q^{(\ell)}, W_v^{(\ell)}, W_o^{(\ell)}, W_{gate}^{(\ell)}, W_{up}^{(\ell)}, W_{down}^{(\ell)}\}_{\ell=1}^{L}$, and matrix $W_{head}$ of head layer.
\State \textbf{Output:} Orthogonally transformed model $\mathcal{M}'$ which is computational invariant to $\mathcal{M}$.
\State Perform SVD over $W_k^{(0)}$, $W_k^{(0)} = U \Sigma V^\top$.
\State $Q_k^{(0)} \leftarrow V.$
\State $\widehat{W}_{emb} \leftarrow W_{emb} Q_k^{(0)}$ 
\For{$\ell = 1, 2, \ldots, L$} 
    \State $\widehat{W}_k^{(\ell)} \leftarrow W_k^{(\ell)} Q_k^{(\ell)}, \quad
    \widehat{W}_q^{(\ell)} \leftarrow W_q^{(\ell)} Q_k^{(\ell)}, \quad
    \widehat{W}_v^{(\ell)} \leftarrow W_v^{(\ell)} Q_k^{(\ell)}$ 
    \State Perform SVD over $W_{gate}^{(\ell)} = U \Sigma V^\top$.
    \State $Q_{gate}^{(\ell)} \leftarrow V$
    \State $\widehat{W}_o^{(\ell)} \leftarrow (Q_{gate}^{(\ell)})^\top W_o^{(\ell)}$
    \State $\widehat{W}_{gate}^{(\ell)} \leftarrow W_{gate}^{(\ell)} Q_{gate}^{(\ell)}, \quad
        \widehat{W}_{up}^{(\ell)} \leftarrow W_{up}^{(\ell)} Q_{gate}^{(\ell)}$
    \If{$\ell < L$} 
        \State Perform SVD over $W_k^{(\ell+1)} = U \Sigma V^\top$ 
        \State $Q_k^{(\ell+1)} \leftarrow V$
        \State $\hat W_{down}^{(\ell)} \leftarrow (Q_k^{(\ell)})^\top W_{down}^{(\ell)}$
    \EndIf    
\EndFor
\end{algorithmic}
\end{algorithm}

\subsection{Results of different models over GSM8K}
\begin{table}[htbp]
  \caption{Results of different models over GSM8K}
  \scriptsize
  \label{tab:gsm8k_results}
  \begin{tabularx}{\textwidth}{lCCCCC}
    \toprule
    Sparsity  & Method & 
    Llama-2-7B & Llama-3-8B & Mistral-7B &
    Phi-4 (14B) \\
    \midrule
     0 & Baseline & 13.95 &	49.96 & 38.74	& 90.22\\
     \midrule
    \multirow{2}{*}{0.25}
    & TEAL  & {14.33} & 48.98 & {37.98} & 89.84 \\
    & \cellcolor{blue!10}\textbf{WINA} & \cellcolor{blue!10}12.51 & \cellcolor{blue!10} {49.36} & \cellcolor{blue!10}37.04 & \cellcolor{blue!10} {90.22}\\\midrule 
    \multirow{2}{*}{0.4}
    & TEAL &  13.12 & 39.88 & 33.97 & 88.02\\
    & \cellcolor{blue!10}\textbf{WINA} &  \cellcolor{blue!10} {13.34} & \cellcolor{blue!10} {40.26} & \cellcolor{blue!10} {35.41} & \cellcolor{blue!10} {90.67}\\\midrule
    \multirow{2}{*}{0.5}
    & TEAL &  8.72 & 27.07 & 29.49 & 86.13\\
    & \cellcolor{blue!10}\textbf{WINA} & \cellcolor{blue!10} {11.22} & \cellcolor{blue!10} {30.40} & \cellcolor{blue!10}{29.72} & \cellcolor{blue!10} {87.57} \\\midrule
    \multirow{2}{*}{0.65}
    & TEAL &  2.50 & 2.73 & 10.99 & 74.37\\
    & \cellcolor{blue!10}\textbf{WINA} & \cellcolor{blue!10} {4.62} & \cellcolor{blue!10} {5.91} & \cellcolor{blue!10} {12.36} & \cellcolor{blue!10} {77.10}\\
    \bottomrule
  \end{tabularx}
\end{table}
%
\subsection{Speed-ups across batch sizes}\label{sec:speedup_batch_size}
\begin{figure}
    \centering
    \includegraphics[width=0.6\linewidth]{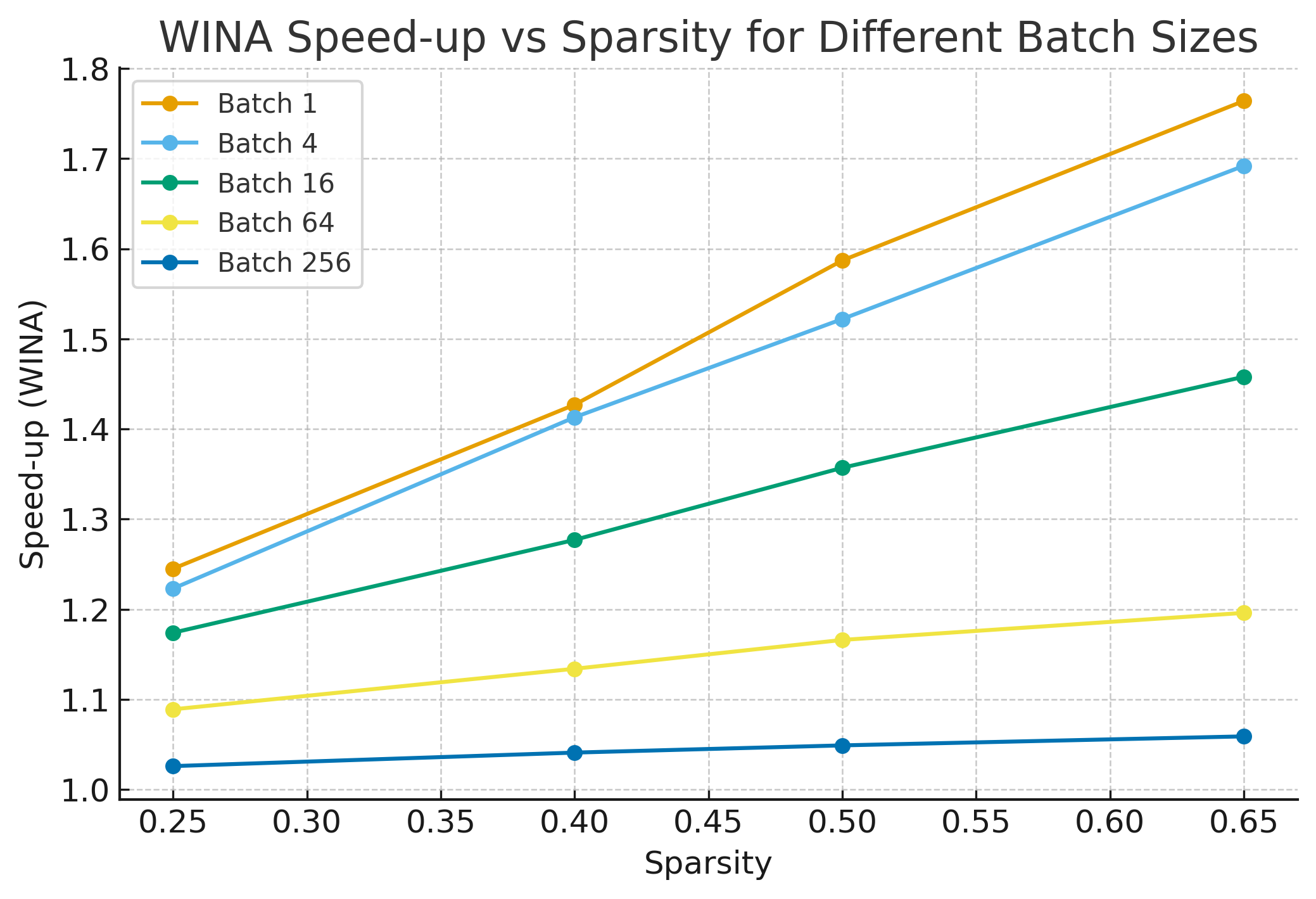}
    \caption{Speed-up ratio versus sparsity level on WINA on $5120\times 17920$.}
    \label{fig:speedup_ratio_batchsize}
\end{figure}

We compute GEMV (General matrix-vector multiplication) latency and speedup via sparse activation across different batch size. The experiment is conducted on A800. We employ a matrix of shape 5120×17920, corresponding to the largest weight matrix in phi-4, the biggest LLM used in our experiments. The sequence length is set to 128. We calculate the speedup based on latency, defined as the ratio of the dense model's latency to the latency under sparse activation. The results in \autoref{fig:speedup_ratio_batchsize} show that TEAL and \algacro{} achieve comparable speedups at the same sparsity level, and the speedup for both decreases as the batch size increases.

\newpage
\subsection{Additional Experiments on LongBench}
\label{subsec: longbench_results}

\begin{table}[htbp]
  \caption{Comparison of WINA and TEAL on LongBench tasks (Llama-3-8B-8k).}
  \scriptsize
  \label{tab:comparison}
  \begin{tabularx}{\textwidth}{lCCCCCCCC}
    \toprule
    Sparsity & Method & 
    Code Completion & Few-shot Learning & Summari-zation &
    Multi-Document QA & Single-Document QA & Synthetic Tasks & Overall \\
    \midrule
     0 & Benchmark &
    23.56 & 60.36 & 16.10 &
    9.87 & 13.80 & 12.76 & 23.14 \\\midrule
    \multirow{2}{*}{0.25}
    & TEAL & 22.00 & {60.17} & 15.77 & 9.96 & {13.92} & 11.03 & 22.69 \\
    & \cellcolor{blue!10}WINA & \cellcolor{blue!10} {23.30} & \cellcolor{blue!10}60.20 & \cellcolor{blue!10}{16.24} & \cellcolor{blue!10}{9.89} & \cellcolor{blue!10}13.65 & \cellcolor{blue!10}{10.92}& \cellcolor{blue!10}\textbf{22.82} \\\midrule
    \multirow{2}{*}{0.4}
    & TEAL & 21.06 & {60.46} & 15.63 & 9.55 & 13.71 & 4.74 & 21.61 \\
    & \cellcolor{blue!10}WINA & \cellcolor{blue!10}{24.13} & \cellcolor{blue!10}60.47 & \cellcolor{blue!10}{17.71} & \cellcolor{blue!10}{9.74} &  \cellcolor{blue!10}{13.02} & \cellcolor{blue!10}{7.30} & \cellcolor{blue!10}\textbf{22.57} \\\midrule
    \multirow{2}{*}{0.5}
    & TEAL & 17.17 & 60.50 & 15.92 & 9.18 & {14.22} & 4.41 & 21.28 \\
    & \cellcolor{blue!10}WINA & \cellcolor{blue!10}{23.88} & \cellcolor{blue!10}{60.24} &  \cellcolor{blue!10}19.29 & \cellcolor{blue!10}9.08 & \cellcolor{blue!10}11.82 & \cellcolor{blue!10}3.95 & \cellcolor{blue!10}\textbf{21.97} \\\midrule
    \multirow{2}{*}{0.65}
    & TEAL & 8.29 & 51.25 & 13.86 & 7.85 & 9.78 & 3.48 & 17.05 \\
    & \cellcolor{blue!10}WINA & \cellcolor{blue!10}{19.58} &  \cellcolor{blue!10}53.61 &  \cellcolor{blue!10}{14.66} & \cellcolor{blue!10}{7.07} & \cellcolor{blue!10}7.73 & \cellcolor{blue!10}3.07 & \cellcolor{blue!10}\textbf{18.13} \\
    \bottomrule
  \end{tabularx}
\end{table}

\begin{table}[htbp]
  \caption{Comparison of WINA and TEAL on LongBench tasks (Phi-4-16k).}
  \scriptsize
  \label{tab:comparison}
  \begin{tabularx}{\textwidth}{lCCCCCCCC}
    \toprule
    Sparsity & Method & 
    Code Completion & Few-shot Learning & Summari-zation &
    Multi-Document QA & Single-Document QA & Synthetic Tasks & Overall \\
    \midrule
     0 & Benchmark & 29.58 & 56.03 & 8.41 & 4.80 & 18.59 & 59.59 & 28.06
     \\\midrule
    \multirow{2}{*}{0.25}
    & TEAL  &  {31.16} & 54.83 &  {10.11} & 5.73 & 18.32 & 55.60 & 27.86\\
    & \cellcolor{blue!10}WINA & \cellcolor{blue!10}30.15 & \cellcolor{blue!10} {55.53} & \cellcolor{blue!10}9.07 & \cellcolor{blue!10} {6.48} & \cellcolor{blue!10} {19.34} & \cellcolor{blue!10} {57.46} & \cellcolor{blue!10}\textbf{28.30}\\\midrule
    \multirow{2}{*}{0.4}
    & TEAL & 30.65 & 55.20 &  {12.53} & 9.95 &  {21.38} & 48.70 & 28.74\\
    & \cellcolor{blue!10}WINA & \cellcolor{blue!10} {33.51} & \cellcolor{blue!10} {56.13} & \cellcolor{blue!10}11.68 & \cellcolor{blue!10} {11.28} & \cellcolor{blue!10}19.60 & \cellcolor{blue!10} {52.10} & \cellcolor{blue!10}\textbf{29.43}\\
    \midrule 
    \multirow{2}{*}{0.5}
    & TEAL & 29.51 & 58.44 & {15.04} & 11.69 &  {22.15} &  {51.03} & 30.54\\
    & \cellcolor{blue!10}WINA & \cellcolor{blue!10}36.53 & \cellcolor{blue!10}{59.39} & \cellcolor{blue!10}13.39 & \cellcolor{blue!10}15.98 & \cellcolor{blue!10}21.19 & \cellcolor{blue!10}48.76 & \cellcolor{blue!10}\textbf{31.39}\\\midrule
    \multirow{2}{*}{0.65}
    & TEAL & 25.71  &  {59.48} & {19.50} & 14.11  & 19.25 & 43.52 & 30.07\\ 
    & \cellcolor{blue!10}WINA & \cellcolor{blue!10} {37.60} & \cellcolor{blue!10}59.22 & \cellcolor{blue!10} {17.36} & \cellcolor{blue!10}12.24 & \cellcolor{blue!10}16.30 & \cellcolor{blue!10} {46.58} & \cellcolor{blue!10}\textbf{30.26} \\
    \bottomrule
  \end{tabularx}
\end{table}

\newpage
\subsection{Extensive Experiments on Social Bias}
\label{sec:social_bias}

Prior work \citep{strubell2023biascompress} has shown that model compression can unintentionally influence social bias in LLMs. To assess whether our sparsification method exhibits similar behavior, we follow the evaluation protocol and stereotype-score metric of \citep{strubell2023biascompress} and measure performance on the CrowS-Pairs benchmark across \textsc{Gender}, \textsc{Race}, and \textsc{Religion} categories. Table \ref{tab:bias-sparsity} reports results for our models at varying sparsity levels.

Across all models, we observe no systematic amplification of social bias as sparsity increases. Instead, bias scores typically change modestly (within 3-6 points) and often move closer to the 50\% ``unbiased'' target defined in \citep{strubell2023biascompress}. 

For instance, Llama-2-7B exhibits reductions in \textsc{Gender} (59.92 $\to$ 53.82) and \textsc{Race} (69.77 $\to$ 66.28) at 65\% sparsity; Llama-3-8B shows mild decreases across all categories at higher sparsity levels; and Mistral-7B maintains largely stable behavior with slight improvements in \textsc{Race} and \textsc{Religion}. Notably, Phi-4-14B demonstrates the strongest effect, with substantial reductions at 65\% sparsity across all categories, particularly in the \textsc{Religion} category (74.29 $\to$ 60.00).

\begin{table}[h!]
\centering
\caption{CrowS-Pairs stereotype scores for Gender, Race, and Religion for \algacro{} over different LLMs. For each metric, we report the stereotype scores and their absolute distance from the optimal unbiased value of 50\% in parenthesis (smaller values indicate less bias from neutrality,). Scores further from 50\% indicate more bias. Green indicates improved bias scores from baseline, red indicates worse, and grey indicates no change from baseline.}
\label{tab:bias-sparsity}
\begin{tabular}{l l l l l}
\toprule
{Model} & {Sparsity} & \textsc{Gender} & \textsc{Race} & \textsc{Religion} \\
\midrule

\multirow{5}{*}{{Llama-2-7B}}
& Baseline & 59.92 (9.92)  & 69.77 (19.77) & 74.29 (24.29) \\
& 25\%     & \cellcolor{gray!20}59.92 (9.92)  & \cellcolor{green!10}67.83 (17.83) & \cellcolor{orange!10}77.14 (27.14) \\
& 40\%     & \cellcolor{green!10}58.40 (8.40)  & \cellcolor{gray!20}71.51 (21.51) & \cellcolor{orange!10}78.10 (28.10) \\
& 50\%     & \cellcolor{green!10}59.54 (9.54)  & \cellcolor{green!10}68.02 (18.02) & \cellcolor{gray!20}74.29 (24.29) \\
& 65\%     & \cellcolor{green!10}53.82 (3.82)  & \cellcolor{green!10}66.28 (16.28) & \cellcolor{orange!10}76.19 (26.19) \\
\midrule

\multirow{5}{*}{{Llama-3-8B}}
& Baseline & 60.31 (10.31) & 66.28 (16.28) & 74.29 (24.29) \\
& 25\%     & \cellcolor{green!10}59.54 (9.54)  & \cellcolor{green!10}65.70 (15.70) & \cellcolor{orange!10}76.19 (26.19) \\
& 40\%     & \cellcolor{orange!10}61.45 (11.45) & \cellcolor{green!10}66.09 (16.09) & \cellcolor{orange!10}76.19 (26.19) \\
& 50\%     & \cellcolor{green!10}57.63 (7.63)  & \cellcolor{green!10}65.50 (15.50) & \cellcolor{green!10}71.43 (21.43) \\
& 65\%     & \cellcolor{orange!10}60.69 (10.69) & \cellcolor{green!10}64.34 (14.34) & \cellcolor{green!10}68.57 (18.57) \\
\midrule

\multirow{5}{*}{{Mistral-7B}}
& Baseline & 62.98 (12.98) & 67.25 (17.25) & 69.52 (19.52) \\
& 25\%     & \cellcolor{green!10}62.60 (12.60) & \cellcolor{orange!10}68.22 (18.22) & \cellcolor{green!10}66.67 (16.67) \\
& 40\%     & \cellcolor{green!10}62.21 (12.21) & \cellcolor{green!10}64.53 (14.53) & \cellcolor{green!10}68.57 (18.57) \\
& 50\%     & \cellcolor{green!10}62.60 (12.60) & \cellcolor{green!10}67.05 (17.05) & \cellcolor{orange!10}69.52 (19.52) \\
& 65\%     & \cellcolor{green!10}61.83 (11.83) & \cellcolor{green!10}64.15 (14.15) & \cellcolor{green!10}64.76 (14.76) \\
\midrule

\multirow{5}{*}{{Phi-4-14B}}
& Baseline & 65.65 (15.65) & 63.95 (13.95) & 74.29 (24.29) \\
& 25\%     & \cellcolor{green!10}63.36 (13.36) & \cellcolor{green!10}63.57 (13.57) & \cellcolor{green!10}71.43 (21.43) \\
& 40\%     & \cellcolor{green!10}61.83 (11.83) & \cellcolor{green!10}60.66 (10.66) & \cellcolor{green!10}67.62 (17.62) \\
& 50\%     & \cellcolor{green!10}59.54 (9.54)  & \cellcolor{green!10}61.43 (11.43) & \cellcolor{green!10}65.71 (15.71) \\
& 65\%     & \cellcolor{green!10}59.54 (9.54)  & \cellcolor{green!10}59.88 (9.88)  & \cellcolor{green!10}60.00 (10.00) \\
\bottomrule
\end{tabular}
\end{table}

Taken together, our results suggest that our sparsification method does not exacerbate social bias and in some cases modestly reduces it, consistent with the behavior of certain compression settings reported in \citep{strubell2023biascompress}: all in all, we see reduced bias in 12/16 experiments for \textsc{Gender}, 14/16 for \textsc{Race}, and 9/16 for \textsc{Religion}. We emphasize, however, that these benchmarks capture only a limited set of bias dimensions. We view these findings as indicative rather than comprehensive, and further research on fairness-aware compression remains an important direction.
\newpage
\subsection{Proof of Lemma \ref{lemma.single_layer}}\label{app:lemma3.1}

\paragraph{Proof.} 

Let $\mathcal{I}^{=0}(\bm{x}):=\{i| \bm{x}_i=0\}$ be the set of indices of zero elements at $\bm{x}$. The output deviation between the original network output and the gated output via a general-format sparsification is:

\begin{align*}
\norm{W(\bm{x}_{\mathcal{I}^{=0}} - \bm{x})}^2 &= \left\| \sum_{i\in \mathcal{I}^{=0}} \bm{x}_i W_{:,i} \right\|_2^2 \\
&= \left( \sum_{i\in \mathcal{I}^{=0}} x_i W_{:,i} \right)^{\top} \left( \sum_{i\in \mathcal{I}^{=0}} \bm{x}_i W_{:,i} \right) \\
&= \sum_{j \in  \mathcal{I}^{=0}} \sum_{i \in  \mathcal{I}^{=0}} \bm{x}_j \bm{x}_i W_{:,j}^\top W_{:,i} \\
&= \sum_{i \in  \mathcal{I}^{=0}} \bm{x}_i^2 \| W_{:,i} \|_2^2 + \sum_{i\neq j  \in  \mathcal{I}^{=0}} \bm{x}_j \bm{x}_i W_{:,j}^\top W_{:,i}
\end{align*}

The expected output deviation for \algacro{} is:



\begin{align*}
e_{\text{\algacro{}}} &= \norm{W\bm{x}_{\mathcal{I}^{=0}_{\text{\algacro{}}}} - W\bm{x}}^2 \\
&= \sum_{i \in \mathcal{I}^{=0}_{\text{\algacro{}}}} \bm{x}_i^2 \| W_{:,i} \|_2^2 + \sum_{i \neq j \in \mathcal{I}^{=0}_{\text{\algacro{}}}} \bm{x}_j \bm{x}_i W_{:,j}^\top W_{:,i}.
\end{align*}

Since $W$ is assumed to be column orthogonal, the cross-term expectations vanish, and the expected output error is determined solely by the main term:
$$
e_{\text{\algacro{}}} = \sum_{i \in \mathcal{I}^{=0}_{\text{\algacro{}}}} \bm{x}_i^2 \| W_{:,i} \|_2^2.
$$

Because \algacro{} sparsification sets the $k$ smallest $|\bm{x}_i\bm{c}_i|$ terms to zero, we have the mask of \algacro{} reaches out the lower bound of approximation error for a single layer network, \ie,
\begin{equation}
\bm{g}_{\text{\algacro{}}}(\bm{x})=\argmin_{\bm{g}\in\{0,1\}^{n}}\quad \norm{W(\bm{x}\odot \bm{g} - \bm{x})}^2.
\end{equation}

Thus, the above indicates that \algacro{} sparsification achieves the tight lower bound of the approximation error, including those of TEAL and CATS. 

\subsection{Proof of Theorem \ref{theorem.L_layer_with_activation}}
\label{app:thm3.5}

\paragraph{Proof.}
We consider an $L$-layer \emph{linear} network (no activation):
\[
\bm{y}^{(1)} = W^{(1)}\bm{x}, 
\qquad
\bm{y}^{(\ell+1)} = W^{(\ell+1)}\bm{y}^{(\ell)}, \ \ell=1,\dots,L-1,
\]
so
\[
\mathcal{M}(\bm{x}) = W^{(L)}\cdots W^{(1)}\bm{x}.
\]

Let $\bm{g}^{(\ell)}\in\{0,1\}^{d_\ell}$ be a gate and define $M^{(\ell)}=\mathrm{diag}(\bm{g}^{(\ell)})$.
The gated network is defined recursively by
\[
\bm{y}_{\bm{g}}^{(\ell)} = W^{(\ell)}\big(M^{(\ell)}\bm{y}_{\bm{g}}^{(\ell-1)}\big),
\qquad
\bm{y}_{\bm{g}}^{(0)}=\bm{x}.
\]
Denote $\mathcal{G}=\{\bm{g}^{(1)},\dots,\bm{g}^{(L)}\}$.

\paragraph{Key assumption.}
For each layer $\ell\ge 2$, assume \emph{column-orthogonality}:
\[
W^{(\ell)\top}W^{(\ell)} = D^{(\ell)} \ \text{is diagonal.}
\]
(Equivalently, columns of $W^{(\ell)}$ are mutually orthogonal; no normalization is required.)

\paragraph{Step 1: Base case $N=2$ and exact decomposition.}
For two layers,
\[
\bm{y}^{(2)} = W^{(2)}W^{(1)}\bm{x},
\qquad
\bm{y}_{\bm{g}}^{(2)} = W^{(2)}M^{(2)}W^{(1)}M^{(1)}\bm{x}.
\]
Let the deviation be $\bm{e}_{\bm{g}}^{(2)} := \bm{y}_{\bm{g}}^{(2)}-\bm{y}^{(2)}$.
Add and subtract $W^{(2)}W^{(1)}M^{(1)}\bm{x}$:
\[
\bm{e}_{\bm{g}}^{(2)}
=
\underbrace{W^{(2)}(M^{(2)}-I)W^{(1)}M^{(1)}\bm{x}}_{:=\bm{v}}
+
\underbrace{W^{(2)}W^{(1)}(M^{(1)}-I)\bm{x}}_{:=\bm{u}}.
\]
Therefore,
\[
\|\bm{e}_{\bm{g}}^{(2)}\|_2^2
=
\|\bm{u}\|_2^2+\|\bm{v}\|_2^2+2\,\bm{u}^\top\bm{v}.
\]
This identity is exact.

\paragraph{Separable expressions for $\|\bm{u}\|_2^2$ and $\|\bm{v}\|_2^2$.}
Let $D^{(2)}:=W^{(2)\top}W^{(2)}$ which is diagonal by assumption.
Write $\bm{u}=W^{(2)}\bm{a}$ and $\bm{v}=W^{(2)}\bm{b}$ where
\[
\bm{a}:=W^{(1)}(M^{(1)}-I)\bm{x},
\qquad
\bm{b}:=(M^{(2)}-I)W^{(1)}M^{(1)}\bm{x}.
\]
Then
\[
\|\bm{u}\|_2^2=\bm{a}^\top D^{(2)}\bm{a},
\qquad
\|\bm{v}\|_2^2=\bm{b}^\top D^{(2)}\bm{b}.
\]
Because $D^{(2)}$ is diagonal and $(M^{(2)}-I)$ keeps exactly the coordinates with $g^{(2)}_j=0$,
\[
\|\bm{v}\|_2^2
=
\sum_{j:g^{(2)}_j=0}\|W^{(2)}_{:,j}\|_2^2 \cdot \big((W^{(1)}M^{(1)}\bm{x})_j\big)^2.
\]
Moreover, letting $A:=W^{(2)}W^{(1)}$ and noting $\bm{u}=A(M^{(1)}-I)\bm{x}$, we have
\[
\|\bm{u}\|_2^2
=
\sum_{i:g^{(1)}_i=0}\|A_{:,i}\|_2^2 \cdot x_i^2.
\]

The cross-term equals
\[
\bm{u}^\top\bm{v}
=
\bm{a}^\top D^{(2)}\bm{b}
=
\sum_{j} D^{(2)}_{jj}\,\bm{a}_j\,\bm{b}_j,
\]
which in general \emph{depends on the gates} and cannot be assumed to vanish without additional structure.

\paragraph{A provable upper bound.}
For any $\alpha>0$, by Young's inequality,
\[
2\,\bm{u}^\top\bm{v}\le \alpha\|\bm{u}\|_2^2+\frac{1}{\alpha}\|\bm{v}\|_2^2,
\]
hence
\[
\|\bm{e}_{\bm{g}}^{(2)}\|_2^2
\le
(1+\alpha)\|\bm{u}\|_2^2+\left(1+\frac{1}{\alpha}\right)\|\bm{v}\|_2^2.
\]
Define the upper bound objective
\[
\mathcal{U}_\alpha(\bm{x};\mathcal{G})
:=
(1+\alpha)\|\bm{u}\|_2^2+\left(1+\frac{1}{\alpha}\right)\|\bm{v}\|_2^2.
\]
Since $\|\bm{u}\|_2^2$ and $\|\bm{v}\|_2^2$ have separable (top-$k$) forms above under column-orthogonality,
WINA's selection rule minimizes $\mathcal{U}_\alpha(\bm{x};\mathcal{G})$.

\paragraph{Step 2: Inductive extension to $N>2$.}
For $N+1$ layers, let the deviation be $\bm{e}_{\bm{g}}^{(N+1)}:=\bm{y}_{\bm{g}}^{(N+1)}-\bm{y}^{(N+1)}$.
Add and subtract $W^{(N+1)}\bm{y}_{\bm{g}}^{(N)}$:
\[
\bm{e}_{\bm{g}}^{(N+1)}
=
\underbrace{W^{(N+1)}(M^{(N+1)}-I)\bm{y}_{\bm{g}}^{(N)}}_{:=\bm{v}}
+
\underbrace{W^{(N+1)}(\bm{y}_{\bm{g}}^{(N)}-\bm{y}^{(N)})}_{:=\bm{u}}.
\]
Thus,
\[
\|\bm{e}_{\bm{g}}^{(N+1)}\|_2^2
=
\|\bm{u}\|_2^2+\|\bm{v}\|_2^2+2\,\bm{u}^\top\bm{v},
\]
and applying Young's inequality yields
\[
\|\bm{e}_{\bm{g}}^{(N+1)}\|_2^2
\le
(1+\alpha)\|\bm{u}\|_2^2+\left(1+\frac{1}{\alpha}\right)\|\bm{v}\|_2^2
=: \mathcal{U}_\alpha(\bm{x};\mathcal{G}).
\]
Under the column-orthogonality assumption for $W^{(N+1)}$, $\|\bm{v}\|_2^2$ is separable:
\[
\|\bm{v}\|_2^2
=
\sum_{j:g^{(N+1)}_j=0}\|W^{(N+1)}_{:,j}\|_2^2\cdot\big(\bm{y}_{\bm{g},j}^{(N)}\big)^2,
\]
so minimizing it reduces to keeping the $k$ largest weighted coordinates.
Moreover, $\|\bm{u}\|_2^2 = \|W^{(N+1)}(\bm{y}_{\bm{g}}^{(N)}-\bm{y}^{(N)})\|_2^2$
is minimized by recursively applying the same rule to the first $N$ layers.
Therefore WINA minimizes the upper bound $\mathcal{U}_\alpha(\bm{x};\mathcal{G})$ for all $L$.

The above establishes that, under column-orthogonality, WINA optimizes a provable upper bound
$\mathcal{U}_\alpha(\bm{x};\mathcal{G})$ on the true deviation $\|\mathcal{M}(\bm{x})-\mathcal{M}(\bm{x}\mid\mathcal{G})\|_2^2$.
In general, without additional assumptions forcing $\bm{u}^\top\bm{v}=0$ (or gate-independent),
one cannot claim exact optimality for the true deviation.
\qed


\subsection{Limitations}
\label{sec:limitations}

To maximize performance gains and ensure realistic acceleration, \algacro{} requires a few additional operations. First, \algacro{} needs the column-wise norms of the weight matrix $W$, which can be pre-computed once during model loading and reused throughout inference. Second, \algacro{} performs an element-wise product between the input and the pre-computed norms; this operation is extremely lightweight and adds negligible overhead.

The performance benefits of \algacro{} are maximized when the weight matrices in LLMs satisfy the column-wise orthogonality condition. To enforce this property, we adopt the efficient one-off orthogonalization transformation proposed in~\citep{ashkboos2024slicegptcompresslargelanguage}. This transformation is applied a single time prior to inference and produces a numerically equivalent LLMs for further usages.

\newpage
\subsection{Additional Results on Hardware Acceleration}\label{sec:addition_hardware_acceleration}

\begin{figure}[h!]
    \centering
    \includegraphics[width=0.9\linewidth]{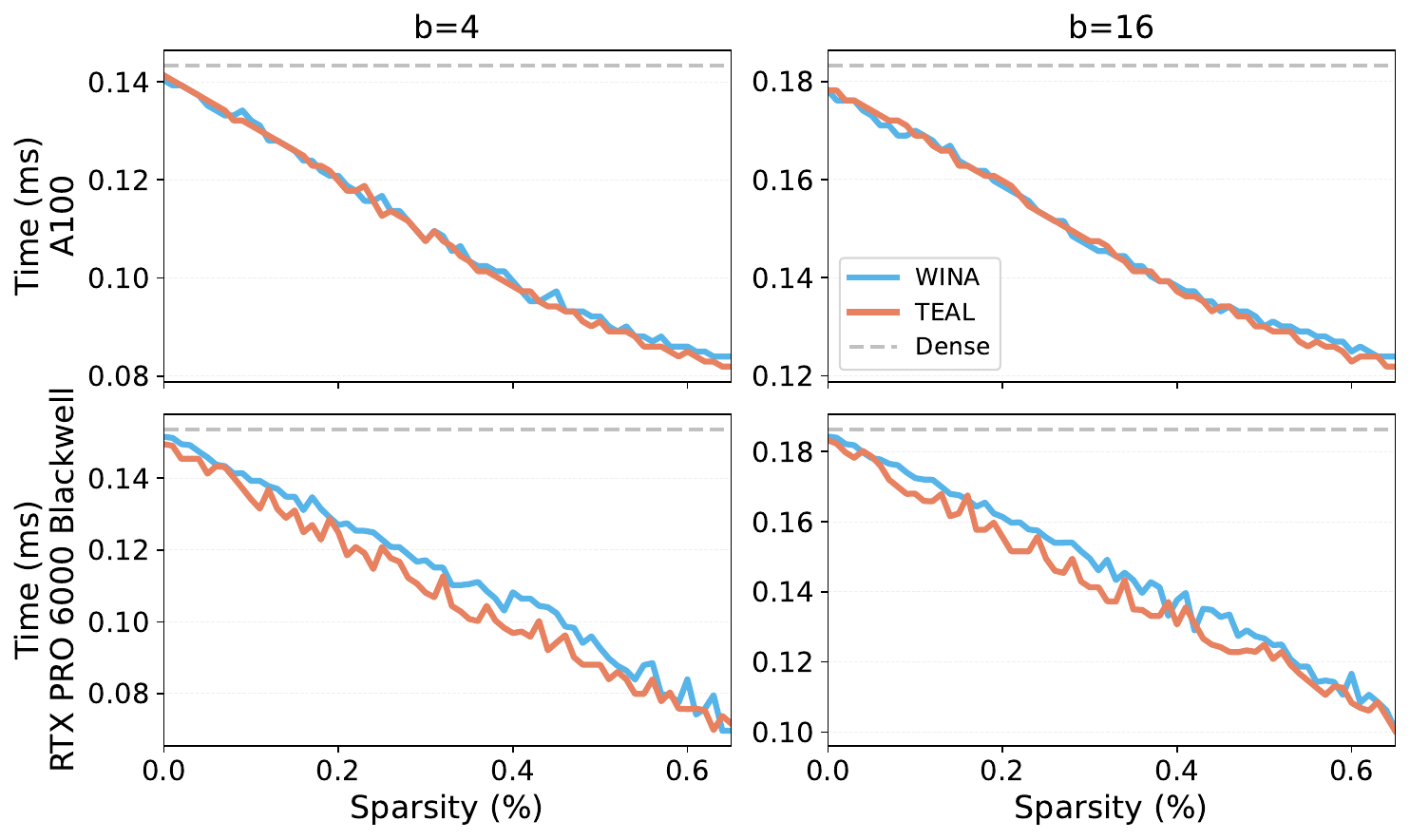}
    \caption{5120x17920 for batch sizes 1 and 4 on A100 and RTX PRO 6000 Blackwell.}
    \label{fig:a100_rtx_5120x17920_bs4_16}
\end{figure}


\begin{figure}[htbp!]
    \centering
    \includegraphics[width=0.9\linewidth]{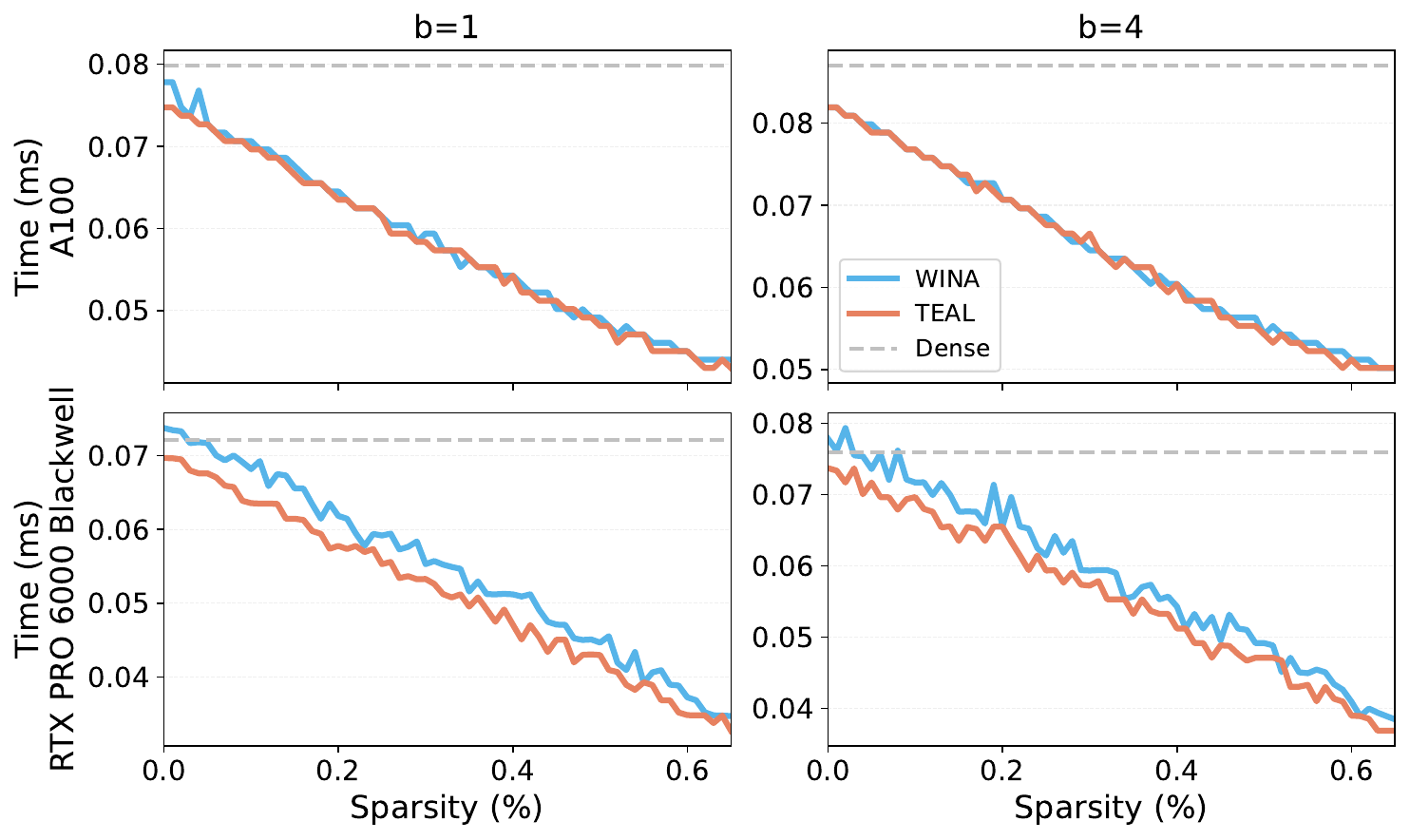}
    \caption{4096x11008 GEMV for batch sizes 1 and 4 on A100 and RTX PRO 6000 Blackwell.}
    \label{fig:a100_rtx_4096x11008_bs14}
\end{figure}








\subsection{Note on Layer-Wise Sparsity Allocation}
Additionally, in this work, we do not focus specifically on per-layer sparsity allocation since \algacro{} is generalized well and applicable over varying sparsity allocations. As future work, we believe dedicated sparsity-allocation strategies and a focus on layer-wise sparsity using different sparsity assignment protocols are promising directions to further improve \algacro{}'s performance. 



\end{document}